\documentclass{ieeeaccess}
\usepackage{cite}
\usepackage{amsmath,amssymb,amsfonts}
\usepackage{algorithmic}
\usepackage{graphicx}
\usepackage{textcomp}
\usepackage{comment}
\usepackage{hyperref}
\usepackage{booktabs}
\usepackage{multirow}
\newcommand{\darr}{\ensuremath{\downarrow}}
\newcommand{\uarr}{\ensuremath{\uparrow}}
\usepackage{textgreek}

\usepackage{bm}
\makeatletter
\AtBeginDocument{\DeclareMathVersion{bold}
\SetSymbolFont{operators}{bold}{T1}{times}{b}{n}
\SetSymbolFont{NewLetters}{bold}{T1}{times}{b}{it}
\SetMathAlphabet{\mathrm}{bold}{T1}{times}{b}{n}
\SetMathAlphabet{\mathit}{bold}{T1}{times}{b}{it}
\SetMathAlphabet{\mathbf}{bold}{T1}{times}{b}{n}
\SetMathAlphabet{\mathtt}{bold}{OT1}{pcr}{b}{n}
\SetSymbolFont{symbols}{bold}{OMS}{cmsy}{b}{n}
\renewcommand\boldmath{\@nomath\boldmath\mathversion{bold}}}
\makeatother

\def\BibTeX{{\rm B\kern-.05em{\sc i\kern-.025em b}\kern-.08em
    T\kern-.1667em\lower.7ex\hbox{E}\kern-.125emX}}

\begin{document}
\history{Date of publication xxxx 00, 0000, date of current version xxxx 00, 0000.}
\doi{10.1109/ACCESS.2024.0429000}

\title{Adaptive Graph Pruning with Sudden-Events Evaluation for Traffic Prediction using Online Semi-Decentralized ST-GNNs}
\author{\uppercase{Ivan Kralj}\authorrefmark{1},
\uppercase{Lodovico Giaretta}\authorrefmark{2},
\uppercase{Gordan Ježić}\authorrefmark{1},
\uppercase{Ivana Podnar Žarko}\authorrefmark{1},
\uppercase{\v{S}ar\=unas Girdzijauskas}\authorrefmark{2}\authorrefmark{3},
}

\address[1]{University of Zagreb, Faculty of Electrical Engineering and Computing, Croatia}
\address[2]{RISE Research Institutes of Sweden, Sweden}
\address[3]{KTH Royal Institute of Technology, Sweden}
%\tfootnote{This paragraph of the first footnote will contain support
%information, including sponsor and financial support acknowledgment. For
%example, ``This work was supported in part by the U.S. Department of
%Commerce under Grant BS123456.''}

\markboth
{I. Kralj \headeretal: Adaptive Graph Pruning with Sudden-Events Evaluation for Traffic Prediction using Online Semi-Decentralized ST-GNNs}
{I. Kralj \headeretal: Adaptive Graph Pruning with Sudden-Events Evaluation for Traffic Prediction using Online Semi-Decentralized ST-GNNs}

\corresp{Corresponding author: Ivan Kralj (e-mail: ivan.kralj@fer.hr).}

\begin{abstract}
Spatio-Temporal Graph Neural Networks (ST-GNNs) are well-suited for processing high-frequency data streams from geographically distributed sensors in smart mobility systems. However, their deployment at the edge across distributed compute nodes (cloudlets) createssubstantial communication overhead due to repeated transmission of overlapping node features between neighbouring cloudlets. To address this, we propose an adaptive pruning algorithm that dynamically filters redundant neighbour features while preserving the most informative spatial context for prediction. The algorithm adjusts pruning rates based on recent model performance, allowing each cloudlet to focus on regions experiencing traffic changes without compromising accuracy. Additionally, we introduce the Sudden Event Prediction Accuracy (SEPA), a novel event-focused metric designed to measure responsiveness to traffic slowdowns and recoveries, which are often missed by standard error metrics. We evaluate our approach in an online semi-decentralized setting with traditional FL, server-free FL, and Gossip Learning on two large-scale traffic datasets, PeMS-BAY and PeMSD7-M, across short-, mid-, and long-term prediction horizons. Experiments show that, in contrast to standard metrics, SEPA exposes the true value of spatial connectivity in predicting dynamic and irregular traffic. Our adaptive pruning algorithm maintains prediction accuracy while significantly lowering communication cost in all online semi-decentralized settings, demonstrating that communication can be reduced without compromising responsiveness to critical traffic events.
\end{abstract}


\titlepgskip=-21pt

\maketitle

\section{Introduction}

Smart mobility has become a cornerstone of modern urban planning and operations, particularly in transportation systems, influencing long-term development and day-to-day traffic management \cite{introduction_1}. It is enabled by the pervasive deployment of smart sensors across the mobility infrastructure, which continuously monitor and report real-time traffic conditions. Within this landscape, traffic prediction is a key component of smart mobility \cite{introduction_2, introduction_3, introduction_4, introduction_5, introduction_6}, encompassing forecasting across different time horizons and tasks such as vehicle speed and public transport forecasting. Effective traffic prediction enables proactive congestion management, dynamic signal control, travel time estimation, and better allocation of resources.

Like most prediction problems, traffic prediction methods can generally be divided into classical statistical methods such as AutoRegressive Integrated Moving Average (ARIMA) \cite{introduction_7} and Machine Learning (ML) methods. The latter can be further split into ``traditional'' ML methods, such as gradient boosting \cite{introduction_8}, and Deep Learning (DL)-based methods including Long Short-Term Memory (LSTM) \cite{introduction_9} and Diffusion Convolutional Recurrent Neural Networks (DCRNN) \cite{introduction_10}. While these methods differ in expressiveness and computational profile, they largely emphasize temporal dependencies and overlook spatial ones.

Graph-based modeling addresses this limitation by representing transportation systems explicitly as graphs, with nodes for physical locations and edges for spatial relations. Graph Neural Networks (GNNs) \cite{preliminaries_4} natively process such structure. Specifically, Spatio-Temporal Graph Neural Networks (ST-GNNs) \cite{preliminaries_11} extend GNNs to capture both spatial and temporal dependencies, making them more effective than other DL-based approaches for traffic forecasting.

However, most deployments and prior studies assume a centralized training and inference pipeline that continuously aggregates measurements from a large sensor field into a single data plane. In practice, collecting data, performing DL training and inference, and sending back commands in real time is difficult to achieve reliably in a centralized system. As the sensor network grows, the central server must scale accordingly to sustain rising data rates and model updates. Moreover, failures or overloads at the core can degrade or interrupt service for the entire network. Additionally, a centralized system can become a significant target for hostile actors and a major cyber-weakness in strategic infrastructure management. These limitations motivate a shift toward solutions that are inherently scalable and resilient at the edge, moving computation closer to data sources, embracing online (real-time) learning and adaptation, and enabling collaboration among geographically distributed compute nodes known as cloudlets \cite{methodology_1}. This cloudlet-based approach forms the foundation for semi-decentralized learning in smart mobility systems.

Previous work on semi-decentralized ST-GNN training across cloudlets \cite{related_work_12} showed that most communication overhead comes from repeatedly exchanging node features between neighbouring cloudlets, resulting in redundant data transfers and inefficient use of bandwidth.

\begin{figure*}[ht!]
  \centering
  \includegraphics[width=1\linewidth]{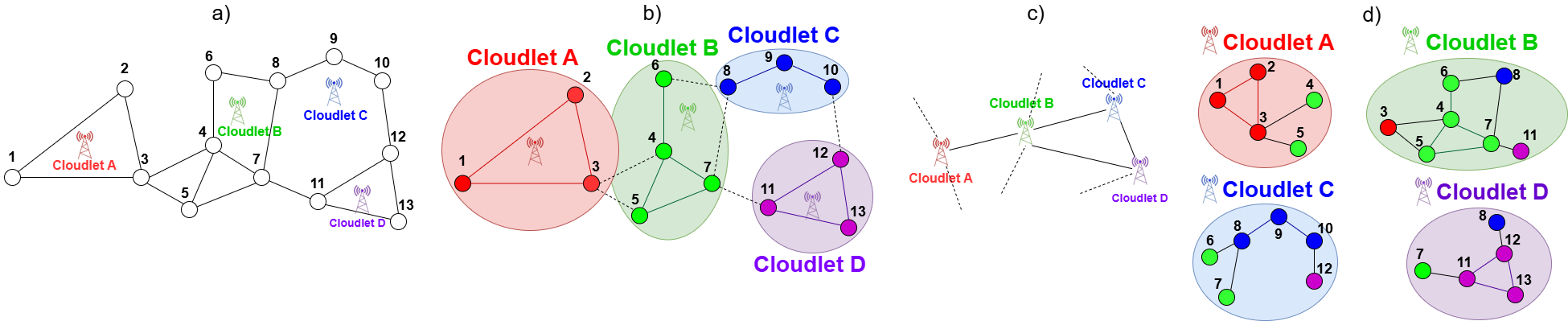}
  \caption{Graph partitioning and communication. a) Geographically distributed sensor network and base stations. b) Graph partitioning of the sensors into cloudlets based on geographical proximity. c) Cloudlet-to-cloudlet communication network for exchanging node features and model updates. d) After communicating with neighbouring cloudlets, each cloudlet can construct the ST-GNN subgraph required for training on its local nodes \cite{related_work_12}.}
  \label{fig:1}
\end{figure*}

To reduce this communication overhead without compromising predictive accuracy, we introduce an adaptive cross-cloudlet pruning algorithm that dynamically limits the amount of features exchanged between neighbouring cloudlets, preserving only the most relevant features originating from nodes located at the edges of neighbouring cloudlets. The pruning aggressiveness is continuously adjusted for each individual cloudlet based on recent validation feedback, allowing the system to reduce communication during stable traffic conditions while retaining essential spatial information in regions experiencing rapid changes.

However, standard evaluation metrics such as MAE, RMSE, and (W)MAPE fail to capture a model's responsiveness to abrupt traffic events, which are critical for real-world traffic management \cite{related_work_15, related_work_16}. While gradual changes in traffic conditions are relatively easy for most forecasting models to track, real systems must also react to sudden slowdowns, i.e., traffic flow breakdown, which arises from complex interactions in traffic dynamics \cite{related_work_20}. However, just as traffic can abruptly shift from free-flow to congested conditions, it can also abruptly recover from congestion to free flow, as seen in our datasets. To address this gap, we introduce a task-specific, event-centric metric, Sudden Event Prediction Accuracy (SEPA), which quantifies how reliably a model detects and predicts sudden slowdowns and recoveries in vehicle speed. Our algorithm uses SEPA to adjust pruning aggressiveness for each individual cloudlet, ensuring that communication is reduced without sacrificing responsiveness to critical traffic events.

We demonstrate the effectiveness of our algorithm within an online semi-decentralized framework in which cloudlets learn from streaming data via sliding windows and synchronize models using different training approaches. The contributions of this paper can be summarized as follows:
\begin{itemize}
    \item We develop an adaptive cross-cloudlet pruning algorithm, a boundary-aware, performance-driven pruning strategy that adaptively removes cross-cloudlet nodes during training without sacrificing model accuracy.

    \item Using oracle-style baselines, we empirically show that models with superior standard evaluation metrics can completely fail to detect sudden traffic slowdowns and recoveries, illustrating that standard metrics are misaligned with the needs of real-world traffic management and motivating our event-centric metric.

    \item Using the findings from above, we introduce Sudden Event Prediction Accuracy (SEPA), which quantifies how reliably models detect slowdowns and recoveries in streaming settings, exposing differences that standard metrics often miss, especially at longer horizons.

    \item We develop a simulation framework for semi-decentralized ST-GNN online training across cloudlets, where cloudlets learn from streaming data via sliding windows and synchronize models under different training approaches.

    \item Using this framework, we extensively evaluate three training approaches---traditional FL, server-free FL, and Gossip Learning---on two real-world datasets across multiple prediction horizons, validating the relevance of SEPA and the effectiveness of our pruning algorithm.
\end{itemize}

Our experiments show that, for all semi-decentralized training approaches, datasets, and prediction horizons, standard validation metrics remain close among all connectivity setups, whereas SEPA shows that full cross-cloudlet connectivity and our adaptive cross-cloudlet pruning algorithm consistently detect more sudden events than the no cross-cloudlet connectivity, especially for long-term prediction. Our pruning algorithm lowers the communication cost of each cloudlet while maintaining competitive predictive accuracy. Finally, cloudlet-level analyses reveal geographical heterogeneity in the online training setup, underscoring the value of cloudlet-aware approaches such as ours. The complete source code and accompanying resources used in this work are accessible through our open GitHub repository\footnote{\label{footnote_0}\url{https://github.com/AIoTwin/dec-stgnn}}.

\section{Preliminaries}\label{sec:preliminares}
\subsection{Traffic prediction}
Traffic prediction is a fundamental time-series forecasting task that estimates traffic metrics such as vehicle speed, volume, and density in order to monitor current road conditions and anticipate future trends. It is generally classified into three forecasting horizons: short-term (15 min), mid-term (30 min), and long-term (60 min) \cite{preliminaries_1}. Accurate forecasting is a key enabler of smart mobility applications, including route optimization, congestion reduction, incident management, and dynamic traffic signal control.

\subsection{Spatio-Temporal Graph Representation of Traffic prediction}
Road networks can be naturally represented as graphs, since roads and intersections inherently form a structured topology. In general, road traffic system is modeled as a spatio-temporal graph at the $t$-th time step, denoted by $\mathcal{G}_t = (\mathcal{V}_t, \mathcal{E}, W)$,
where $\mathcal{V}_t = \{v_1, v_2, \dots, v_n\}$ is the finite set of nodes, each corresponding $n$ IoT-devices that captures real-time traffic observations, $(v_i, v_j)\in \mathcal{E}$ is the set of spatial edges connecting the IoT-devices, and $W \in \mathbb{R}^{n \times n}$ is the weighted adjacency matrix, where $W_{i,j}$ encodes the spatial dependency between these IoT-devices.

The spatial component is derived from the physical distance between IoT-devices, while the temporal component is introduced through the sequence of observed traffic features evolving over time. At each node $v_i$, the observation of IoT-device $u$ at time $t$ is denoted as $X_u^{t} \in \mathbb{R}^d$, where $d$ represents the number of traffic metrics and $X^{t} \in \mathbb{R}^{|V| \times d}$ represents the traffic metrics of all IoT-devices at time $t$ \cite{preliminaries_2}.

The aim of traffic prediction is to learn a function $f$ that takes $T$ historical observations as input to predict the next $T'$ traffic metrics in the future from $n$ correlated IoT-devices on the road network \cite{preliminaries_3}.
\begin{equation}
    [X_\mathcal{G}^{(t-T+1)}, \dots, X_\mathcal{G}^{t}] \xrightarrow[]{\text{f}} [X_\mathcal{G}^{t+1}, \dots, X_\mathcal{G}^{t+T'}]
\end{equation}

\subsection{Graph Neural Networks}
Graph Neural Networks (GNNs) \cite{preliminaries_4} are deep neural networks that are designed to operate on graph-structured data by iteratively exchanging and transforming information along edges. Any graph can be represented as $G = (V, E)$, where $V =\{v_1, \dots ,v_{|V|}\}$ is set of nodes denoted as $v_i \in V$, and $E = \{e_1, \dots, e_{|V|}\}$ is set of edges connecting the nodes denoted as $e_{ij} = (v_i,v_j) \in E$. In machine learning, graph are typically defined as $G = (X, A)$, where $A \in \mathbb{R}^{|V| \times |V|}$ is adjacency matrix, where $|V|$ represents total number of nodes, with $A_{ij} = 1$ if $e_{ij} \in E$ and $A_{ij} = 0$ if $e_{ij} \notin E$, and $X \in \mathbb{R}^{|V| \times M}$ is a node feature matrix, where $|V|$ represents total number of nodes, and $M$ represents total number of node features.

Like other neural networks, GNNs consist of $l$-layers, where each layer transforms node features by aggregating information from neighbours. The output of each layer is a new set of node embeddings $H \in \mathbb{R}^{|V| \times D}$, which capture increasingly larger neighbourhoods as the network deepens, where $|V|$ represents total number of nodes, and $D$ is a hyperparameter typically defined by a user to determine the vector's length.

GNNs are categorized into Convolutional Graph Neural Networks (ConvGNNs) and Recurrent Graph Neural Networks (RecGNNs). ConvGNNs extend the idea of convolution to graphs, focusing on aggregating features from spatially connected neighbours, such as Graph Convolutional Network (GCN) \cite{preliminaries_5}, Chebyshev Graph Convolutional Neural Networks (ChebConv) \cite{preliminaries_6} and Graph Sample and Aggregation (GraphSAGE) \cite{preliminaries_7}. RecGNNs apply recurrent updates over the graph structure, performing iterative message passing until node embeddings converge, such as $GNN^{*2}$ by Scarselli et al.\cite{preliminaries_8}, Graph Echo State Network (GraphESN) \cite{preliminaries_9} and Gated Graph Neural Network (GGNN) \cite{preliminaries_10}.

While ConvGNNs are capable of capturing spatial dependencies among road sensors, they do not account for the temporal evolution of traffic signals, which is essential in traffic prediction. Similarly, RecGNNs, despite their iterative message passing scheme, also operate on static graphs and model structural dependencies until convergence, rather than temporal dynamics. Therefore, neither ConvGNNs nor RecGNNs are suitable for our problem setting, which requires joint modeling of both spatial and temporal dependencies.

\subsection{Spatio-Temporal Graph Neural Networks}
Spatio-Temporal Graph Neural Networks (ST-GNNs) \cite{preliminaries_11} are DL models that extend GNNs to model both spatial and temporal dependencies in structured data, allowing it to simultaneously account for spatial correlations between nodes and temporal correlations over time, making it particularly suited for traffic prediction. Spatio-temporal graph can be represented as $\mathcal{G}_t = (\mathcal{V}_t, \mathcal{E}_t, A_t)$ where $\mathcal{V}$ is a finite set of nodes, $\mathcal{E}_t$ is a set of edges, and $A_t$ denotes the adjacency matrix at time $t$ \cite{preliminaries_16}. ST-GNNs are designed to handle both spatial and temporal dependencies simultaneously, making them particularly well-suited for traffic forecasting \cite{preliminaries_12}.

A number of ST-GNN architectures have been developed \cite{preliminaries_11}, differing in how they couple spatial and temporal modules, such as Diffusion Convolutional Recurrent Neural Network (DCRNN) \cite{preliminaries_13}, Spatio-Temporal Graph Convolutional Network (ST-GCN) \cite{preliminaries_14}, Graph Attention LSTM Network \cite{preliminaries_15}, and many more \cite{preliminaries_11}. In this work, we utilize ST-GCN due to their superior ability to handle traffic prediction tasks in the IoT context compared to other techniques. Experiments demonstrated that ST-GCN consistently outperforms other models in key evaluation metrics (MAE, MAPE, and RMSE) \cite{preliminaries_14}.

\subsection{Distributed Training Paradigms}
In distributed machine learning, the placement of data and computation, the coordination mechanism, and what is exchanged among participants define the training paradigm. We categorize distributed training based on their system ar- chitecture and coordination mechanisms:

\begin{itemize}
    \item \textbf{Centralized training:} Data is collected and processed in a single location, which maintains and updates a global model \cite{preliminaries_22}. This setup simplifies coordination and often yields strong performance, but creates a single point of failure and raises scalability, privacy, and regulatory concerns \cite{preliminaries_23, preliminaries_24}.
    
    \item \textbf{Semi-decentralized training:} A hybrid approach, which combines centralized and decentralized schemes, where devices are organized into discrete cluster or hierarchies. Local coordination occurs within clusters as model updates are sent to higher-level aggregators \cite{preliminaries_25}. By distributing aggregation across multiple cluster heads, semi-decentralized training reduces the communication burden on a single server and improves scalability as the number of participating devices grows, but the introduction of a hierarchical structure increases system complexity and it still remains vulnerable to imbalance or failures at cluster heads, which can affect fairness and robustness across the network \cite{preliminaries_26}.
    
    \item \textbf{Fully-decentralized training:} Devices exchange model updates among other participating devices in a peer-to-peer manner. Through iterative rounds of local updates and neighbour-to-neighbour exchanges, models gradually converge to a consensus without requiring any global aggregation \cite{preliminaries_22, preliminaries_27}. This removes single points of failure and bottlenecks, and can naturally scale to large and dynamics networks. However, without a central or hierarchical aggregator, convergence can be slower and less stable. In addition, the total number of communication rounds required to reach consensus may be significantly higher compared to centralized or semi-decentralized approaches \cite{related_work_12}.
\end{itemize}

% It previously used preliminaries_18 as well, check section_v1 for more details
\paragraph{\textbf{Traditional federated learning.}}
Traditional FL \cite{preliminaries_17, preliminaries_18} can be seen as a semi-decentralized scheme where multiple clients work together to train a ML model under the coordination of a central aggregator, while keeping their raw data stored locally. Clients perform local training using their private data and only share model parameters or gradients with the server. The central server then aggregates these updates to form an improved global model, which is redistributed to the clients for the next training round. This process iterates until the global model converges.

\paragraph{\textbf{Server-free federated learning.}}
Server-free FL \cite{preliminaries_19} is a decentralized variant of the traditional FL paradigm that removes the reliance on a central aggregator. Instead of a single coordinating entity that aggregates client updates, the learning process is distributed entirely across participating clients. Each client performs training using its private data and periodically exchanges model updates with its neighbours in the network topology. At the same time, each client receives updates from its neighbours and incorporates these contributions into its own model. This neighbour-to-neighbour communication continues over multiple rounds, allowing knowledge to propagate through the network until all nodes gradually converge to a consensus global model.

% It previously used preliminaries_21 as well, check section_v1 for more details
\paragraph{\textbf{Gossip learning.}}
Gossip Learning \cite{preliminaries_20, preliminaries_21} is a fully-decentralized protocol for training ML models without the need for a central coordinator. Instead, models are sent through the network in the form of random walks. Each device stores two models in its memory (maintained in a FIFO buffer), typically representing the most recent models received from neighbouring devices. At each iteration of the gossip protocol, a device averages the weights of the two most recent models in its FIFO buffer to create an aggregated model. The device then performs one local training step using its own private data, refining the aggregated model based on its local observations. The updated model is then forwarded to a randomly selected peer in the network. This process is repeated across the network, ensuring that models progressively evolve and improve as they traverse different devices, integrating knowledge from different data distributions at each location.

\subsection{Online training}
In online training, the model is updated continuously as new data arrives, rather than being trained once on a fixed dataset \cite{preliminaries_28}. This paradigm enables models to adapt to evolving environments by incorporating streaming data into the learning process, making it well-suited for dynamic and non-stationary domains. Unlike offline or batch learning, which assumes access to the entire dataset before training, online training operates incrementally and learns in real time.

The training process is iterative. At each step, the learner receives a new sample or a small batch of samples, updates its parameters, and then proceeds to the next data point. To remain efficient, the learner typically retains a small buffer of past samples, ensuring that the model does not overfit to only the most recent data \cite{preliminaries_29}.
\section{Methodology}
\subsection{Scenario}
The traffic prediction scenario involves deployment of IoT-devices along highways. These IoT-devices are installed at fixed locations to continuously capture traffic metrics at regular time intervals. To streamline data processing, IoT-devices are assigned to local base stations (BS), also referred to as cloudlets, based on their physical proximity and communication range, ensuring that each IoT-device transmits its reading to exactly one cloudlet. Communication between IoT-devices and their assigned cloudlets is facilitated using Low-Power Wide-Area Network (LPWAN) protocols, such as LoRa or NB-IoT, which provide long-range, energy-efficient, and reliable data transmission suitable for resource-constrained sensors. Cloudlets, in addition to serving as communication hubs, are provisioned with moderate computational resources, enabling them to locally train machine learning models, such as ST-GNNs.

By design, ST-GNNs aggregate information from $\ell$-hop neighbourhoods at each layer, meaning that the representation of a node often depends on traffic features collected from devices located outside the boundaries of its cloudlet. As a result, to correctly compute node embeddings, each cloudlet must obtain additional information from neighbouring cloudlets. Without such cross-cloudlet feature sharing, the local model would suffer from incomplete neighbourhood context and reduced predictive accuracy (as demonstrated in Table~\ref{table:4}). Therefore, cloudlets act both as learners and as collaborators, exchanging only the necessary information to enable accurate distributed training of ST-GNNs.

\subsection{Online Semi-Decentralized ST-GNN Training}
In our online semi-decentralized training scenario, each cloudlet is responsible for training a local ST-GNN model on the data collected from its assigned IoT-devices. Training proceeds in an online fashion: at the beginning of the process, each cloudlet trains on an initial batch of data to establish a baseline model. Afterwards, new traffic observations arrive continuously, and training is performed using a sliding window strategy, where each time window provides the most recent data slice. In each window, the cloudlet updates its local model on the newly arrived data.

We applied this scenario under two different setups. The first serves as a baseline, in which no communication reduction is applied. In this setup, each cloudlet constructs its local ST-GNN subgraph, retrieves all node features from neighbouring cloudlets, and proceeds with training. The second setup integrates our proposed communication-efficient strategy, where cross-cloudlet information exchange is selectively reduced. Instead of transmitting all boundary node features, cloudlets apply an adaptive mechanism to identify and exclude less informative cross-cloudlet nodes from the training process.

\subsection{Distributed Training Setups}
We adopt a semi-decentralized ST-GNN architecture inspired by Nazzal et al. \cite{methodology_1} and refined by Kralj et al. \cite{related_work_12}. To illustrate the key components of such architecture, we provide an overview in Fig.~\ref{fig:1}. In Fig.~\ref{fig:1}-a, traffic network is represented as a graph, where nodes correspond to IoT-devices deployed along highways and edges denote road connections based on physical proximity. In Fig.~\ref{fig:1}-b, the global graph is partitioned into subgraphs, each managed by a local cloudlet based on geographical proximity. In Fig.~\ref{fig:1}-c, the cloudlets form a communication network, enabling them to exchange the necessary node features and model updates with their neighbours. Finally, in Fig.~\ref{fig:1}-d, each cloudlet leverages this exchanged information to construct the local ST-GNN subgraph required to train on its assigned nodes.

Building on this architecture, we evaluate three different training setups---traditional FL, server-free FL, and Gossip Learning---for ST-GNNs in the context of traffic prediction. These training setups define the baseline against which we compare our proposed algorithm.

\subsection{GNN Partitioning in Distributed Training}
Partitioning the global traffic graph into smaller subgraphs for local training in distributed GNN scenarios introduces unique challenges due to their reliance on spatial dependencies. Unlike traditional ML models, where each device processes localized data independently, GNNs require features from neighbours to compute effective representations. Formally, in an $\ell$-layer GNN, the representation of a node depends not only on its own features but also by those of all nodes within $\ell$-hops of its position in the graph.

This spatial dependency introduces a need for inter-cloudlet communication, as the required node features often extend beyond the boundaries of a single cloudlet's subgraph. Since IoT-devices typically have limited communication capabilities and cannot directly exchange data with distant cloudlets, the responsibility for exchanging node features falls on the cloudlets themselves. Furthermore, as the number of layers grows, the receptive field expands accordingly, causing substantial overlap between neighbouring subgraphs. The result is that the same node features are frequently requested and transmitted multiple times by different cloudlets.

To address these dependencies, each cloudlet maintains awareness of which neighbours require their node features for local computations based on the underlying sensor network connectivity. When new features are collected from the IoT-sensors, each cloudlet proactively broadcasts these features to its neighbouring cloudlets that need them for training (Fig.~\ref{fig:1}-d). This proactive exchange of node features ensures that every cloudlet has the required information needed to perform local computations without compromising the accuracy of the GNN model.

\subsection{Sudden Event Prediction Accuracy (SEPA) Metric}\label{methodology:sape}
The Sudden Event Prediction Accuracy (SEPA) is a task-specific metric designed to evaluate how reliably a model captures abrupt traffic events such as congestion and its recovery. Unlike standard regression metrics that average prediction errors over all time steps, SEPA quantifies how reliably the model captures sudden drops or recoveries in vehicle speed.

Using ground truth, a sudden event is detected for a node at time step $\text{t}$ if, within a recent time window of $\text{H}$ observations, there exists an earlier point $\text{k} < \text{t}$ such that the observed speed difference exceeds a predefined threshold $\delta_{\text{change}}$. Formally, two types of events are considered:
\begin{itemize}
    \item \textbf{Traffic jam (slowdown):} when the current observed speed drops sharply compared to a past value, $\text{X}_\text{t} \leq \text{X}_\text{k} - \delta_{\text{change}}$,
    \item \textbf{Recovery (speed-up):} when the current observed speed rises sharply compared to a past value, $\text{X}_\text{t} \geq \text{X}_\text{k} + \delta_{\text{change}}$.
\end{itemize}

Once a sudden event is detected, we evaluate whether model correctly predicted sudden event by comparing it with the ground truth. A prediction is considered correct if the absolute prediction error at time $\text{t}$ falls within a tolerance margin $\delta_{\text{tol}}$:
\begin{equation}
    | \hat{X}_t - X_t | \leq \delta_{\text{tol}}.
\end{equation}
To avoid double-counting temporally adjacent events, a cooldown period $\tau_{\text{c}}$ is applied after each detection, during which no new events are recorded for that node.

The SEPA for a single node is computed as the fraction of correctly predicted sudden events over the total number of detected events:
\begin{equation}
    \text{SEPA} = \frac{\text{\# correctly predicted sudden events}}{\text{\# total sudden events}}.
\end{equation}

At the cloudlet level, SEPA values are aggregated across all local nodes to produce an overall event-level accuracy rate $\text{SEPA}_\text{t}$ for the current time window $\text{t}$, reflecting how effectively a model detects sudden traffic changes across the spatial region managed by a cloudlet.

\subsection{Adaptive Cross-Cloudlet Pruning Algorithm} %before it was Node Score Algorithm
\begin{figure*}[ht!]
  \centering
  \includegraphics[width=1\linewidth]{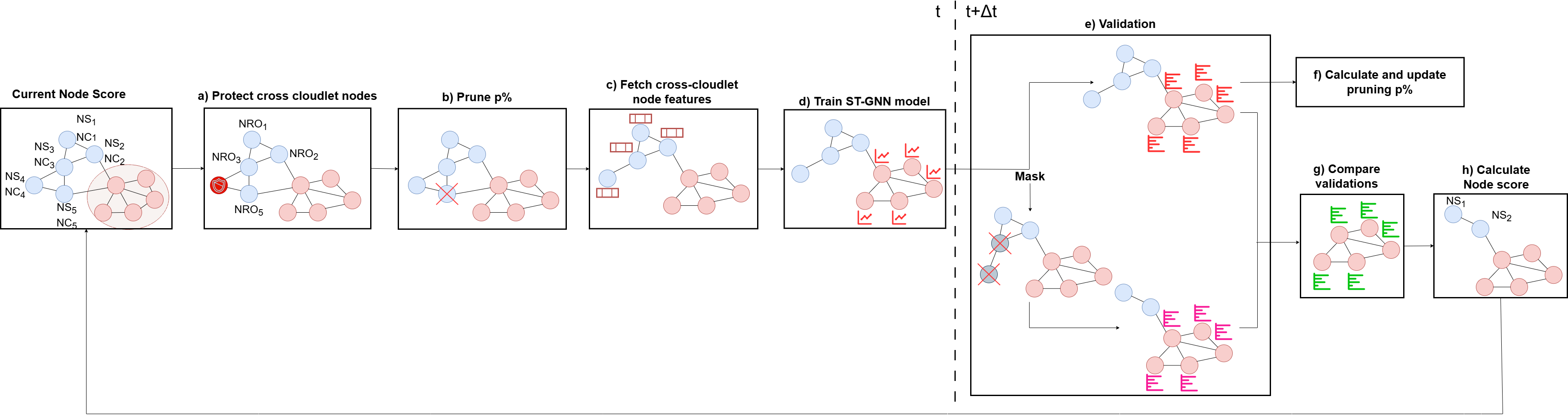}
  \caption{High-level flow of the Adaptive Cross-Cloudlet Pruning Algorithm for online semi-decentralized ST-GNN training}
  \label{fig:2}
\end{figure*}

As established in Fig.~\ref{fig:1}-d, distributed ST-GNN training incurs substantial overhead from repeatedly exchanging node features across cloudlets. Our proposed Adaptive Cross-Cloudlet Pruning Algorithm addresses this by selectively and dynamically pruning cross-cloudlet nodes based on predictive performance and node importance, while preserving the boundary information for future time window. This is done for each individual cloudlet. Figure~\ref{fig:2} illustrates the high-level flow of the algorithm across consecutive time windows.

The first step focuses on local traffic dynamics, as shown in Fig.~\ref{fig:2}-a. Each cloudlet first identifies local nodes experiencing sudden traffic changes, as described in Section \ref{methodology:sape}. If such an event is detected at a local node, we mark its neighbours as protected so they cannot be pruned in the current time window. This safeguard ensures that areas experiencing dynamic changes retain full context from surrounding nodes, preserving the model's ability to react to evolving traffic conditions.

Next, the algorithm prunes a fraction $\text{p}_\text{t}$ of the remaining cross-cloudlet nodes, as shown in Fig.~\ref{fig:2}-b. The pruning follows a probabilistic distribution based on each cross-cloudlet node's score $\text{NS}_\text{i}^\text{t}$, where cross-cloudlet nodes with higher scores (less important) have higher removal probability:
\begin{equation}
    \Pr(\text{prune node } i) \propto NS_i^t, \quad i \in \mathcal{N}_{\text{cross-cloudlet}}.
\end{equation}

Once pruning of cross-cloudlet nodes is completed, each cloudlet constructs a new ST-GNN subgraph, allowing each cloudlet to fetch both local and non-pruned cross-cloudlet nodes features, as shown in Fig.~\ref{fig:2}-c. In Fig.~\ref{fig:2}-d, this new subgraph is used to train the ST-GNN model for the current time window $\text{t}$. 

% How to say that, when we receive data for current timewindow, we train on the "old" data, and the validate on the "current" data
After training, each cloudlet validates its model using data received from current timewindow, as shown in Fig.~\ref{fig:2}-e. The model is first evaluated on the pruned subgraph, and then on a masked variant in which half of the remaining cross-cloudlet nodes are randomly omitted. Both evaluations produce SEPA, denoted as $\text{SEPA}_\text{t}$.

The pruning controller (Fig.~\ref{fig:2}-f) adaptively updates pruning percentage $\text{p}_\text{t}$ using recent validation outcomes. Every $E_{\text{settle}}$ time windows, the SEPA scores from the last $\text{W}$ windows are aggregated and compared with a baseline SEPA obtained during the $\text{W}_{\text{init}}$ time windows of the warm-up phase of the online algorithm. The aggregated baseline rate is computed as:
\begin{equation}
    SEPA_{\text{base}} = \frac{1}{W_{\text{init}}} \sum_{t=0}^{W_{init}} SEPA_t,
\end{equation}
while the adaptive windowed aggregation is updated after every $\text{E}_\text{settle}$ epochs using a FIFO buffer of length W:
\begin{equation}
    SEPA_{\text{win}} = \frac{1}{W} \sum_{t=W_{init}+1-W+1}^{T} SEPA_t.
\end{equation}
The ratio between the recent and baseline aggregation provides a signal for adjusting pruning intensity:
\begin{equation}
    \text{Ratio}_t = \frac{SEPA_{\text{win}}}{SEPA_{\text{base}}}.
\end{equation}

If $\text{Ratio}_t > 1 + \delta_{\text{margin\_up}}$, then pruning is increased by $\delta_{\text{pruning\_up}}$. If $\text{Ratio}_t < 1 - \delta_{\text{margin\_down}}$, then pruning is decreased by $\delta_{\text{margin\_down}}$, otherwise pruning percentage remains unchanged. Formally,
\begin{equation}
p_{t+1} =
\begin{cases}
p_t + \delta_{\text{pruning\_up}}, & \text{if } \text{Ratio}_t > 1 + \delta_{\text{margin\_up}},\\[4pt]
p_t - \delta_{\text{pruning\_down}}, & \text{if } \text{Ratio}_t < 1 - \delta_{\text{margin\_down}},\\[4pt]
p_t, & \text{otherwise}.
\end{cases}
\end{equation}

Here, $\delta_{\text{margin\_up}}$ and $\delta_{\text{margin\_down}}$ represent adaptive performance margins, and $\delta_{\text{pruning\_up}}$, $\Delta_{\text{pruning\_down}}$ are the step sizes controlling the rate of pruning change. Dynamic bounds $[\text{p}_{\min}, \text{p}_{\max}]$ ensure that pruning never exceeds safety limits.

Next, we compute the difference in validation performance between the pruned and masked ST-GNN subgraphs using their SEPA values, as shown in Fig.~\ref{fig:2}-g. Since SEPA increases with better predictive accuracy (higher is better), we first express it in an error-oriented form, where higher values indicate worse performance. The performance degradation associated with masking is then calculated as:
\begin{equation}
    \Delta SEPA^{t} = SEPA_{\text{masked}}^{t} - SEPA_{\text{pruned}}^{t}.
\end{equation}
A negative $\Delta \text{SEPA}^\text{t}$ therefore implies that removing additional nodes (in the masked variant) decreased accuracy, meaning those nodes were important. 

Finally, in Fig.~\ref{fig:2}-h, this feedback is translated into updated node scores. Node score accumulates this information across timewindows, giving higher scores to nodes whose absence consistently reduces accuracy:
\begin{equation}
    NS^{t+1} = \sum_{\tau=0}^{t-1} NS^\tau + \Delta SEPA^{t},
\end{equation}
In this way, each cloudlet maintains a cumulative importance profile of its cross-cloudlet nodes, guiding future pruning decisions in subsequent time windows.

% Add new subsection here regarding "oracle" model (with this, we actually prove that standard eval metrics dont tell whole story)
\subsection{Illustrative Oracle Baselines for Metric Analysis}
To clearly demonstrate why standard regression metrics can fail to reflect behaviour that is critical for real-world traffic management, we construct two illustrative oracle-style predictors. Both are defined directly on the ground-truth speed time series using the same sudden-event detector, but they are deliberately biased in opposite ways in order to show that lower errors doesn't necessarily mean that the model is better for a specific task.

The first baseline model, denoted as event-blind oracle, is constructed to perform perfectly during normal traffic flow, but fails to detect any sudden events. Concretely, for all non-sudden-event time steps the prediction is the same as the ground truth, yielding no errors. However, at time steps corresponding to sudden slowdowns or recoveries, the predicted value is shifted so that the absolute error exceeds the tolerance band $\delta_{\text{tol}}$ used in the SEPA definition. As a result, this model achieves excellent MAE, RMSE, and WMAPE, yet its SEPA score is zero because it never correctly predicts the endpoints of sudden events.

The second baseline model, denoted as event-perfect oracle, is constructed to predict all sudden events correctly, but is intentionally less accurate during normal flow. The predicted value is generated by adding a random value that is smaller than the tolerance band $\delta_{\text{tol}}$ used in the SEPA definition across the entire dataset. As a result, this model achieves a perfect SEPA score, but its MAE, RMSE, and WMAPE are noticeably worse than those of the event-blind oracle.

Table~\ref{table:oracle} reports the resulting metrics for these two baselines using PeMS-BAY dataset, computed over the same evaluation split as our main experiments. This controlled experiment confirms that models optimized or evaluated solely using standard metrics can appear superior despite being ineffective at capturing abrupt traffic changes, whereas an event-centric metric such as SEPA correctly ranks a model as more useful for real-world traffic management.

% TABLE: Oracle baselines
\begin{table}[htbp]
    \centering
    \caption{Performance comparison (MAE [$mile/h$] / RMSE [$mile/h$] / WMAPE [\%] / Sudden Event Prediction Accuracy, SEPA [\%]) for oracle-style baselines using PeMS-BAY dataset.}
    \label{table:oracle}
    \renewcommand{\arraystretch}{1.1}
    \setlength\tabcolsep{6pt}
    \scriptsize
    \begin{tabular}{l c c c c}
        \toprule
        \textbf{Model} & \textbf{MAE} & \textbf{RMSE} & \textbf{WMAPE} & \textbf{SEPA} \\
        \midrule
        Event-blind oracle  & 0.09 & 0.99 & 0.001 & 0.0 \\
        Event-perfect oracle & 1.53 & 1.77 & 0.025 & 100.0 \\
        \bottomrule
    \end{tabular}
\end{table}
\section{Experimental Setup}
\subsection{Datasets}
To evaluate the performance of our algorithm, we utilize two real-world traffic datasets collected by the California Freeway Performance Measurement System (PeMS), PeMS-BAY\footnote{\label{footnote_1}Download link: \url{https://github.com/liyaguang/DCRNN}} \cite{preliminaries_13} and PeMSD7-M\footnote{\label{footnote_2}Download link: \url{https://github.com/VeritasYin/STGCN_IJCAI-18}} \cite{preliminaries_14}. PeMSD7-M collects traffic data from 228 sensor stations in the California state highway system during the weekdays from May through June in 2012, while PeMS-BAY collects traffic data from 325 sensors in the Bay Area of California, starting from January 1st 2017 through May 31th 2017. The details of datasets are listed in Table~\ref{table:1}.

% TABLE I
\begin{table}[htbp]
    \centering
    \caption{Details of PeMS-BAY and PeMSD7-M}
    \begin{tabular}{c c c c c}
        \hline
        Datasets & Nodes & Interval & TimeSteps & Attribute \\
        \hline
        PeMS-BAY & 325 & 5 min & 52,116 & Traffic speed \\
        PeMSD7-M & 228 & 5 min & 12,672 & Traffic speed \\
        \hline
    \end{tabular}
    \label{table:1}
\end{table}

The spatial adjacency matrix is constructed from the actual road network based on distance between sensors. Following the ChebNet formulation \cite{experimental_setup_1}, edge weights are assigned as a decreasing function of distance, ensuring that closer sensors exert stronger influence in the graph structure. Additionally, the data were preprocessed for efficient model training and evaluation. For example, if we want to utilize the historical data spanning one hour to predict the data in the next hour, we pack the sequence data in group of 12 and convert it into an instance. Both datasets were split into training and validation sets with a 80:20 ratio. Unlike offline training, where the validation set is used after training each epoch, our online training setup performs validation using the training dataset, as well as validation dataset. At each iteration, the model is trained on the data from the current window $\text{t}$ and immediately validated on the subsequent window $\text{t + 1}$. This validation strategy enables the model to adapt incrementally to evolving traffic patterns while still being evaluated in a realistic forward-looking manner. Once all training windows are processed, the model is finally evaluated on the validation set, which serves as the final evaluation of the model. Additionally, to mitigate the impact of varying data magnitudes, we applied standardization to normalize the original data values, reducing the impact of the large difference in value.

\subsection{Evaluation Metrics}
In our experiments, we evaluate the performance of our distributed ST-GNN training setups using three standard regression metrics: Mean Absolute Error (MAE), Root Mean Square Error (RMSE), and Weighted Mean Absolute Percentage Error (WMAPE).

\paragraph{Mean Absolute Error (MAE).}
MAE measures the average absolute difference between the predicted and actual values, offering a straightforward measure of prediction accuracy:
\begin{equation}
    \text{MAE}(\bm{x}, \bm{\hat{x}}) = \frac{1}{N} \sum_{i=1}^{N} \lvert x_i - \hat{x}_i \rvert,
\end{equation}

\paragraph{Root Mean Square Error (RMSE).}
RMSE emphasizes larger errors by squaring the deviations before averaging, making it particularly sensitive to outliers and providing insights into the error magnitude:
\begin{equation}
    \text{RMSE}(\bm{x}, \bm{\hat{x}}) = \sqrt{\frac{1}{N} \sum_{i=1}^{N} (x_i - \hat{x}_i)^2}.
\end{equation}

\paragraph{Weighted Mean Absolute Percentage Error (WMAPE).}
To account for differences in vehicle speed magnitudes across regions, we use WMAPE:
\begin{equation}
    \text{WMAPE}(\bm{x}, \bm{\hat{x}}) = \frac{\sum_{i=1}^{N} \lvert x_i - \hat{x}_i \rvert}{\sum_{i=1}^{N} x_i} \times 100\%.
\end{equation}

\noindent where $\bm{x} = \{x_1, \ldots, x_n\}$ denotes the ground truth vehicle speed, $\bm{\hat{x}} = \{\hat{x}_1, \ldots, \hat{x}_n\}$ represents their predicted values, and $N$ is the total number of samples.

All metrics are computed after rescaling predictions back to the original data range, ensuring comparability with raw vehicle speeds. We use WMAPE because it accounts for the varying magnitudes of vehicle speeds across regions, providing a more balanced evaluation, compared to MAPE.

We compute these metrics across three forecasting horizons—--short-, mid-, and long-term—--to capture fluctuations in performance over time. For all training setups, the reported metrics are calculated as a weighted average across the individual cloudlets to ensure a consistent and fair comparison. In addition to these established metrics, we also report SEPA. The event-detection procedure and how SEPA drives pruning are detailed in Methodology section.

% Figure 3
\begin{figure}[h]
  \centering
  \includegraphics[width=1\linewidth]{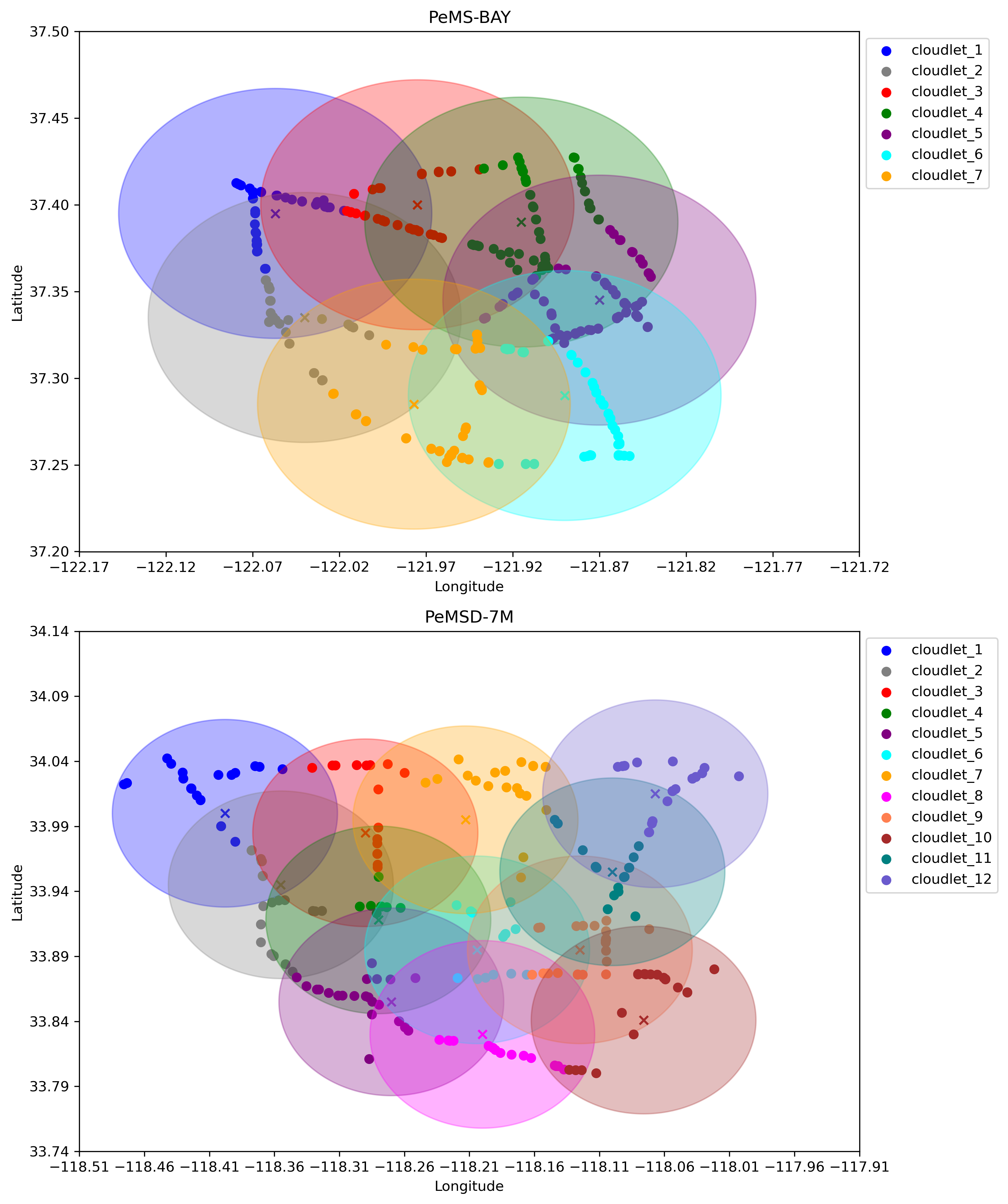}
  \caption{Sensor assignment to cloudlets based on communication range}
  \label{fig:3}
\end{figure}

\subsection{Experimental Setup}
% TABLE II
\begin{table}[h!]
\centering
\caption{Hyperparameters used in the adaptive cross-cloudlet pruning algorithm and SEPA metric}
\label{table:2}
\renewcommand{\arraystretch}{1.1}
\setlength\tabcolsep{6pt}
\scriptsize
\begin{tabular}{l l l}
\toprule
\textbf{Parameters} & \textbf{Symbol} & \textbf{Value} \\
\midrule

\multirow{11}{*}{Pruning controller}
 & $p_{\text{start}}$               & $0.10$ \\
 & $\text{p}_{\min}$                & $0.10$ \\
 & $\text{p}_{\max}$                & $0.70$ \\
 & $\text{W}_{\text{init}}$         & $2$ \\
 & $\text{W}$                       & $3$ \\
 & $\delta_{\text{margin\_up}}$     & $0.00$ \\
 & $\delta_{\text{margin\_down}}$   & $0.03$ \\
 & $\delta_{\text{pruning\_up}}$   & $0.05$ \\
 & $\delta_{\text{pruning\_down}}$ & $0.05$ \\
 & $\text{E}_{\text{settle}}$       & $3$ \\

\midrule

\multirow{4}{*}{SEPA metric}
 & $\text{H}$                       & $12$ \\
 & $\delta_{\text{change}}$         & $20.0$ mile/h \\
 & $\delta_{\text{tol}}$            & $10.0$ mile/h \\
 & $\tau_{\text{c}}$                & $6$ \\
\bottomrule
\end{tabular}
\end{table}

All experiments were executed on the High-Performance Computing (HPC) system Vrančić provided by Srce (University of Zagreb University Computing Centre), which features 4x NVIDIA A100 40GB GPUs, with training assigned to a single GPU. The software stack consisted of Python 3.10.14, PyTorch 2.3.1, and PyTorch Geometric 2.5.3.

For the ST-GNN architecture, we used two spatio-temporal blocks (ST-blocks) with Gated Linear Unit (GLU) activation. The model was trained using the Adam optimizer with Mean Absolute Error (MAE) as the loss function. The learning rate was initialized at 0.0001, with weight decay of $1\mathrm{e}{-5}$, dropout of 0.5, and batch size of 32. A StepLR scheduler was applied, where the learning rate was decayed by a factor of 0.7 every 5 epochs. The spatial kernel size was set to 3, the temporal kernel size was set to 3, and Chebyshev convolution was used for spatial graph filtering. Historical input length was set to 60 minutes (12 data points at 5-minute intervals), with forecasting horizons of 15, 30, and 60 minutes. The hyperparameters used in our experiments are based on previously established configurations that have been optimized for the given model architecture and datasets \cite{preliminaries_14}. We adopt these settings to ensure comparability with prior work and maintain consistency in evaluating different training approaches.

Unlike conventional fixed-epoch training, our online learning framework progresses in sliding windows of data. Each "epoch" corresponds to a predefined amount of newly arrived data, meaning the total number of epochs is determined by the dataset length rather than being fixed in advance. In our experiments, we set data sample amount at 70 and 140 per epoch. Running experiments under both conditions allows us to investigate how the size of the arrival window influences cross-cloudlet communication and predictive accuracy.

Sensors were distributed across multiple cloudlets to simulate a semi-decentralized environment. For the PeMS-BAY dataset, sensors were partitioned into 7 cloudlets, while for PeMSD7-M, 12 cloudlets were used to ensure complete coverage of the geographical area. Cloudlet placement was determined manually based on spatial proximity and communication range, with each cloudlet covering all IoT devices within an 8 km radius. This limitation directly impacts the server-free FL approach by restricting the number of cloudlets that can exchange model updates. In contrast, this constraint does not affect Gossip Learning, as model updates are sent to a randomly selected cloudlet across the entire network, regardless of proximity. Fig.~\ref{fig:3} shows how sensors are assigned to cloudlets and their respective communication ranges for both the PeMS-BAY and PeMSD7-M datasets. This partitioning reflects the semi-decentralized structure used for our distributed training approach.

% We could move it early on????
Our experimental setup is designed to investigate communication-efficient semi-decentralized ST-GNN online training for traffic prediction. Specifically, we evaluate whether the proposed adaptive cross-cloudlet pruning algorithm can substantially reduce cross-cloudlet communication overhead while maintaining predictive accuracy across different training setups. Table~\ref{table:2} lists the fixed controller hyperparameter values used during all experiments. These parameters define how aggressively pruning adapts to recent SEPA performance, and how we detect sudden events in traffic.
\begin{figure*}[ht!]
  \centering
  \includegraphics[width=1\linewidth]{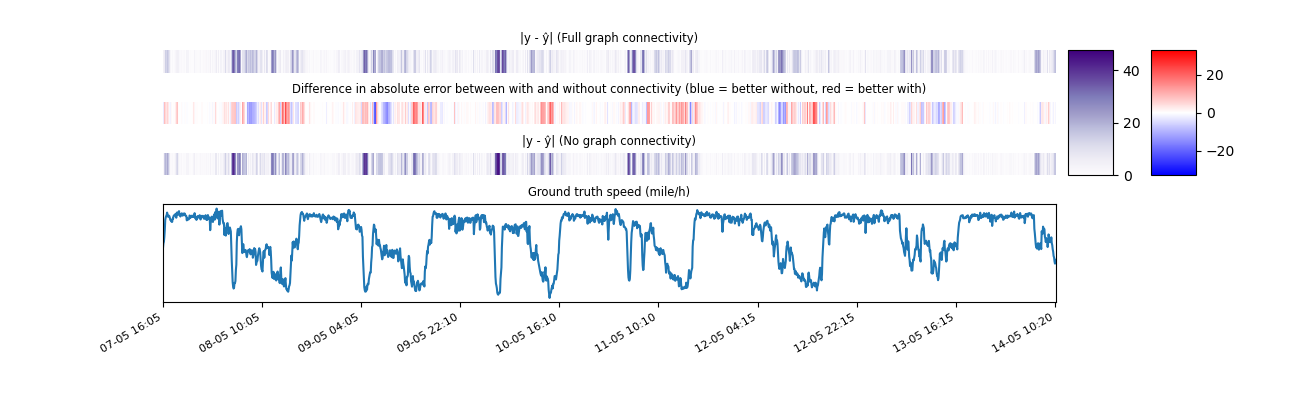}
  \caption{Difference in absolute error between full and no graph connectivity for long-term prediction on the PeMS-BAY dataset (sensor 134)}
  \label{fig:4}
\end{figure*}

% TABLE III
\begin{table}[ht!]
\centering
\caption{Comparison of average absolute error differences (|\textDelta|) between \textit{Full} and \textit{No} cross-cloudlet connectivity, computed over the validation set subsets across all nodes (Long-term prediction).}
\label{table:3}
\setlength\tabcolsep{6pt}
\renewcommand{\arraystretch}{1.25}
\scriptsize
\resizebox{\columnwidth}{!}{%
\begin{tabular}{l c c c c c}
\toprule
Dataset &
\shortstack{Avg $|\Delta|$\\All [mile/h]} &
\shortstack{Avg $|\Delta|$\\Sudden [mile/h]} &
\shortstack{Avg $|\Delta|$\\Non-sudden [mile/h]} &
\shortstack{\#\\Sudden} &
\shortstack{\#\\Non-sudden} \\
\midrule
PeMS-BAY  & 0,33 & 2,44 & 0,32 & 8,188 & 2,524,537 \\
PeMSD7-M  & 0,49 & 2,04 & 0,48 & 2,508 & 425,220 \\
\bottomrule
\end{tabular}
}% end resizebox
\\[2pt]
% \raggedright\footnotesize
% $|\Delta|$ denotes the absolute difference in error between \textit{Full} and \textit{No} graph connectivity, averaged over the indicated subset of the validation set (long-term prediction only).
\end{table}

% TABLE IV
\begin{table*}[ht!]
\centering
\caption{Performance comparison (MAE [$mile/h$] / RMSE [$mile/h$] / WMAPE [\%] / Sudden Event Prediction Accuracy, SEPA [\%])
across multiple different setups, graph connectivity, and timewindow sizes.}
\label{table:4}
\setlength\tabcolsep{3pt}
\renewcommand{\arraystretch}{1.25}
\scriptsize
\resizebox{\textwidth}{!}{%
\begin{tabular}{l l l c *{12}{c}}
\toprule
\multirow{3}{*}{Dataset} &
\multirow{3}{*}{Setup} &
\multirow{3}{*}{Connectivity} &
\multirow{3}{*}{Timewindow} &
\multicolumn{4}{c}{15 min} &
\multicolumn{4}{c}{30 min} &
\multicolumn{4}{c}{60 min} \\
\cmidrule(lr){5-8}\cmidrule(lr){9-12}\cmidrule(lr){13-16}
&&&& MAE\darr & RMSE\darr & WMAPE\darr & SEPA\uarr & MAE\darr & RMSE\darr & WMAPE\darr & SEPA\uarr & MAE\darr & RMSE\darr & WMAPE\darr & SEPA\uarr \\
\midrule

% ====================== PeMS-BAY ======================
\multirow{18}{*}{PeMS-BAY}
  & \multirow{6}{*}{Traditional FL}
    & \multirow{2}{*}{Full cross-cloudlet conn.}      & 70  & 1.53 & 3.20 & 2.46 & 56.43 & 2.10 & 4.44 & 3.36 & 45.67 & \underline{2.72} & 5.66 & 4.35 & 33.26 \\
  & &                                   & 140 & 1.53 & \underline{3.15} & 2.44 & \underline{56.62} & \textbf{2.03} & \textbf{4.36} & \textbf{3.25} & \underline{46.47} & \textbf{2.70} & \textbf{5.61} & \underline{4.33} & \textbf{34.04} \\
  & & \multirow{2}{*}{No cross-cloudlet conn.}    & 70  & 1.53 & 3.19 & 2.44 & 55.32 & 2.08 & 4.49 & 3.34 & 43.44 & 2.84 & 5.87 & 4.56 & 27.45 \\
  & &                                   & 140 & \underline{1.52} & 3.16 & \underline{2.43} & 55.96 & \underline{2.04} & 4.43 & 3.27 & 43.86 & 2.78 & 5.77 & 4.46 & 29.07 \\
  & & \multirow{2}{*}{Adaptive conn. (Ours)}   & 70  & 1.54 & 3.21 & 2.47 & 56.01 & 2.13 & 4.47 & 3.41 & 44.76 & 2.76 & 5.70 & 4.41 & 32.58 \\
  & &                                   & 140 & \textbf{1.51} & \textbf{3.14} & \textbf{2.42} & \textbf{56.96} & \underline{2.04} & \underline{4.37} & \underline{3.26} & \textbf{46.52} & \textbf{2.70} & \underline{5.62} & \textbf{4.32} & \underline{33.30} \\
  
\cmidrule(lr){2-16}
  & \multirow{6}{*}{\shortstack[l]{Server-free\\FL}}
    & \multirow{2}{*}{Full cross-cloudlet conn.}      & 70  & 1.55 & 3.21 & 2.48 & 55.86 & 2.08 & 4.40 & 3.33 & 44.53 & 2.76 & 5.72 & \underline{4.42} & 32.36 \\
  & &                                   & 140 & \underline{1.51} & \textbf{3.14} & \underline{2.41} & \underline{56.84} & \textbf{2.04} & \textbf{4.36} & \textbf{3.27} & \textbf{46.80} & \underline{2.75} & \textbf{5.60} & \textbf{4.40} & \underline{34.11} \\
  & & \multirow{2}{*}{No cross-cloudlet conn.}    & 70  & 1.55 & 3.20 & 2.48 & 55.40 & 2.09 & 4.48 & 3.34 & 43.73 & 2.82 & 5.85 & 4.52 & 29.60 \\
  & &                                   & 140 & \underline{1.51} & \underline{3.16} & 2.42 & 56.19 & \underline{2.05} & 4.41 & \underline{3.28} & 45.12 & \underline{2.75} & 5.73 & \textbf{4.40} & 29.92 \\
  & & \multirow{2}{*}{Adaptive conn. (Ours)}   & 70  & 1.55 & 3.22 & 2.48 & 55.98 & 2.08 & 4.43 & 3.33 & 45.24 & 2.76 & 5.73 & 4.43 & 32.24 \\
  & &                                   & 140 & \textbf{1.50} & \textbf{3.14} & \textbf{2.40} & \textbf{56.88} & \textbf{2.04} & \underline{4.37} & \textbf{3.27} & \underline{45.88} & \textbf{2.74} & \underline{5.61} & \textbf{4.40} & \textbf{34.22} \\
  
\cmidrule(lr){2-16}
  & \multirow{6}{*}{Gossip Learning}
    & \multirow{2}{*}{Full cross-cloudlet conn.}      & 70  & 1.57 & \underline{3.22} & 2.52 & \underline{55.72} & 2.12 & 4.51 & 3.40 & \underline{44.05} & \underline{2.79} & \underline{5.73} & \underline{4.47} & \underline{31.58} \\
  & &                                   & 140 & \underline{1.55} & 3.23 & \underline{2.49} & 55.70 & 2.10 & \textbf{4.43} & 3.36 & 44.30 & \textbf{2.75} & \textbf{5.64} & \textbf{4.41} & \textbf{32.84} \\
  & & \multirow{2}{*}{No cross-cloudlet conn.}    & 70  & \underline{1.55} & 3.24 & \underline{2.49} & 55.25 & 2.09 & 4.51 & 3.34 & 42.90 & 2.85 & 5.90 & 4.57 & 26.96 \\
  & &                                   & 140 & \textbf{1.54} & \textbf{3.21} & \textbf{2.46} & 55.52 & \textbf{2.06} & \underline{4.47} & \textbf{3.30} & 43.73 & 2.81 & 5.87 & 4.50 & 28.11 \\
  & & \multirow{2}{*}{Adaptive conn. (Ours)}   & 70  & 1.70 & 3.45 & 2.73 & 54.27 & 2.16 & 4.62 & 3.46 & 42.90 & 2.91 & 5.98 & 4.66 & 30.98 \\
  & &                                   & 140 & 1.58 & 3.28 & 2.53 & \textbf{55.73} & \underline{2.08} & \textbf{4.43} & \underline{3.33} & \textbf{44.95} & 2.83 & 5.86 & 4.54 & 28.32 \\
\midrule

% ====================== PeMSD7-M ======================
\multirow{18}{*}{PeMSD7-M}
  & \multirow{6}{*}{Traditional FL}
    & \multirow{2}{*}{Full cross-cloudlet conn.}      & 70  & 2.51 & 4.48 & 4.34 & 54.75 & 3.42 & 6.19 & 5.93 & 41.82 & 4.67 & 8.10 & 8.09 & 28.80 \\
  & &                                   & 140 & 2.46 & \textbf{4.37} & 4.26 & \textbf{55.50} & \underline{3.40} & \textbf{6.09} & \underline{5.88} & \textbf{43.08} & \textbf{4.56} & \textbf{7.81} & \textbf{7.91} & \underline{30.84} \\
  & & \multirow{2}{*}{No cross-cloudlet conn.}    & 70  & 2.56 & 4.54 & 4.43 & 53.31 & 3.50 & 6.35 & 6.06 & 37.88 & 4.79 & 8.28 & 8.30 & 24.25 \\
  & &                                   & 140 & \underline{2.45} & 4.42 & \underline{4.24} & 53.88 & 3.42 & 6.17 & 5.92 & 39.55 & 4.65 & 8.09 & 8.06 & 25.84 \\
  & & \multirow{2}{*}{Adaptive conn. (Ours)}   & 70  & 2.60 & 4.60 & 4.49 & 54.05 & 3.45 & 6.20 & 5.97 & 41.68 & 4.70 & 8.20 & 8.14 & \textbf{31.11} \\
  & &                                   & 140 & \textbf{2.44} & \underline{4.39} & \textbf{4.23} & \underline{54.78} & \textbf{3.37} & \underline{6.10} & \textbf{5.84} & \underline{42.81} & \underline{4.63} & \underline{7.95} & \underline{8.02} & 29.09 \\
  
\cmidrule(lr){2-16}
  & \multirow{6}{*}{\shortstack[l]{Server-free\\FL}}
    & \multirow{2}{*}{Full cross-cloudlet conn.}      & 70  & 2.59 & 4.62 & 4.84 & 54.10 & 3.50 & 6.25 & 6.06 & 41.36 & \underline{4.58} & 8.10 & \underline{7.95} & \underline{31.13} \\
  & &                                   & 140 & \textbf{2.44} & \textbf{4.38} & \textbf{4.23} & \textbf{55.71} & \textbf{3.37} & \textbf{6.05} & \textbf{5.84} & \underline{42.89} & \textbf{4.54} & \textbf{7.81} & \textbf{7.87} & \textbf{31.65} \\
  & & \multirow{2}{*}{No cross-cloudlet conn.}    & 70  & 2.56 & 4.63 & 4.44 & 53.45 & 3.45 & 6.32 & 5.98 & 38.27 & 4.79 & 8.34 & 8.30 & 24.09 \\
  & &                                   & 140 & \underline{2.46} & 4.42 & \underline{4.26} & 54.37 & \underline{3.38} & 6.15 & 5.86 & 41.58 & 4.64 & 8.09 & 8.05 & 26.56 \\
  & & \multirow{2}{*}{Adaptive conn. (Ours)}   & 70  & 2.56 & 4.59 & 4.43 & 54.09 & 3.45 & 6.19 & 5.98 & 42.32 & 4.74 & 8.12 & 8.22 & 30.37 \\
  & &                                   & 140 & \textbf{2.44} & \underline{4.39} & \textbf{4.23} & \underline{55.14} & 3.39 & \underline{6.06} & \underline{5.87} & \textbf{43.57} & 4.59 & \underline{7.90} & 7.96 & 30.16 \\
  
\cmidrule(lr){2-16}
  & \multirow{6}{*}{Gossip Learning}
    & \multirow{2}{*}{Full cross-cloudlet conn.}      & 70  & 2.70 & 4.76 & 4.68 & 53.11 & 3.74 & 6.63 & 6.48 & 38.07 & \underline{4.73} & 8.18 & \underline{8.20} & 29.25 \\
  & &                                   & 140 & \textbf{2.50} & \textbf{4.44} & \textbf{4.33} & \textbf{55.19} & \underline{3.48} & \underline{6.17} & \underline{6.04} & \underline{41.56} & \textbf{4.61} & \textbf{8.01} & \textbf{7.99} & \textbf{32.45} \\
  & & \multirow{2}{*}{No cross-cloudlet conn.}    & 70  & 2.77 & 4.89 & 4.81 & 51.66 & 3.58 & 6.46 & 6.20 & 36.41 & 4.92 & 8.42 & 8.54 & 24.34 \\
  & &                                   & 140 & \underline{2.56} & \underline{4.55} & \underline{4.44} & 53.74 & 3.50 & 6.33 & 6.06 & 38.71 & 4.76 & 8.19 & 8.26 & 26.51 \\
  & & \multirow{2}{*}{Adaptive conn. (Ours)}   & 70  & 2.89 & 5.01 & 5.01 & 51.86 & 3.65 & 6.40 & 6.33 & \textbf{41.78} & 4.93 & 8.40 & 8.54 & 29.17 \\
  & &                                   & 140 & \underline{2.56} & 4.56 & \underline{4.44} & \underline{54.26} & \textbf{3.44} & \textbf{6.16} & \textbf{5.97} & 41.50 & \underline{4.73} & \underline{8.09} & 8.21 & \underline{30.23} \\
\bottomrule
\end{tabular}
}% end resizebox
\end{table*}

\section{Experimental Results}
\subsection{Analysis of node-level impact with graph connectivity pruning}
To better understand the role of graph connectivity at the node level, we conducted an experiment using the PeMS-BAY dataset with long-term prediction. Two models were trained in a centralized offline setting: one without any changes to the nodes edge connectivity (i.e., full graph connectivity) and one without any edges between nodes (i.e., no graph connectivity). For visualization, a single node (sensor 134) and certain period were selected, as plotting all nodes and full dataset would make the figure unreadable. The results are presented in Fig.~\ref{fig:4}, which contains four subplots: (i) the absolute difference between ground truth and prediction for the full graph connectivity, (ii) the difference in absolute error between full and no graph connectivity, (iii) the absolute difference between ground truth and prediction for the no graph connectivity, and (iv) the ground-truth speed across the entire validation set.

From these plots, it is not immediately clear which setup is superior. There are periods where the full graph model outperforms, but also periods where the no-graph model achieves lower difference error compared to the ground truth value. However, when focusing specifically on points when vehicle speed changes suddenly--—such as the onset of congestion or recovery after a traffic jam--—the advantage of full graph connectivity becomes evident. As summarized in Table~\ref{table:3}, when averaging absolute errors only over time steps corresponding to sudden changes, the error difference between the full and no graph configurations increases substantially in favor of full graph connectivity. In contrast, during periods of stable traffic flow, this difference remains minimal.

This observation supports the introduction of SEPA, our new evaluation metric, which directly measures a model's ability to capture these critical events. Additionally, it supports the use of our adaptive cross-cloudlet pruning algorithm, which aims to reduce redundant cross-cloudlet feature sharing while ensuring that nodes most relevant for sudden changes are preserved.

\begin{figure*}[ht!]
  \centering
  \includegraphics[width=1\linewidth]{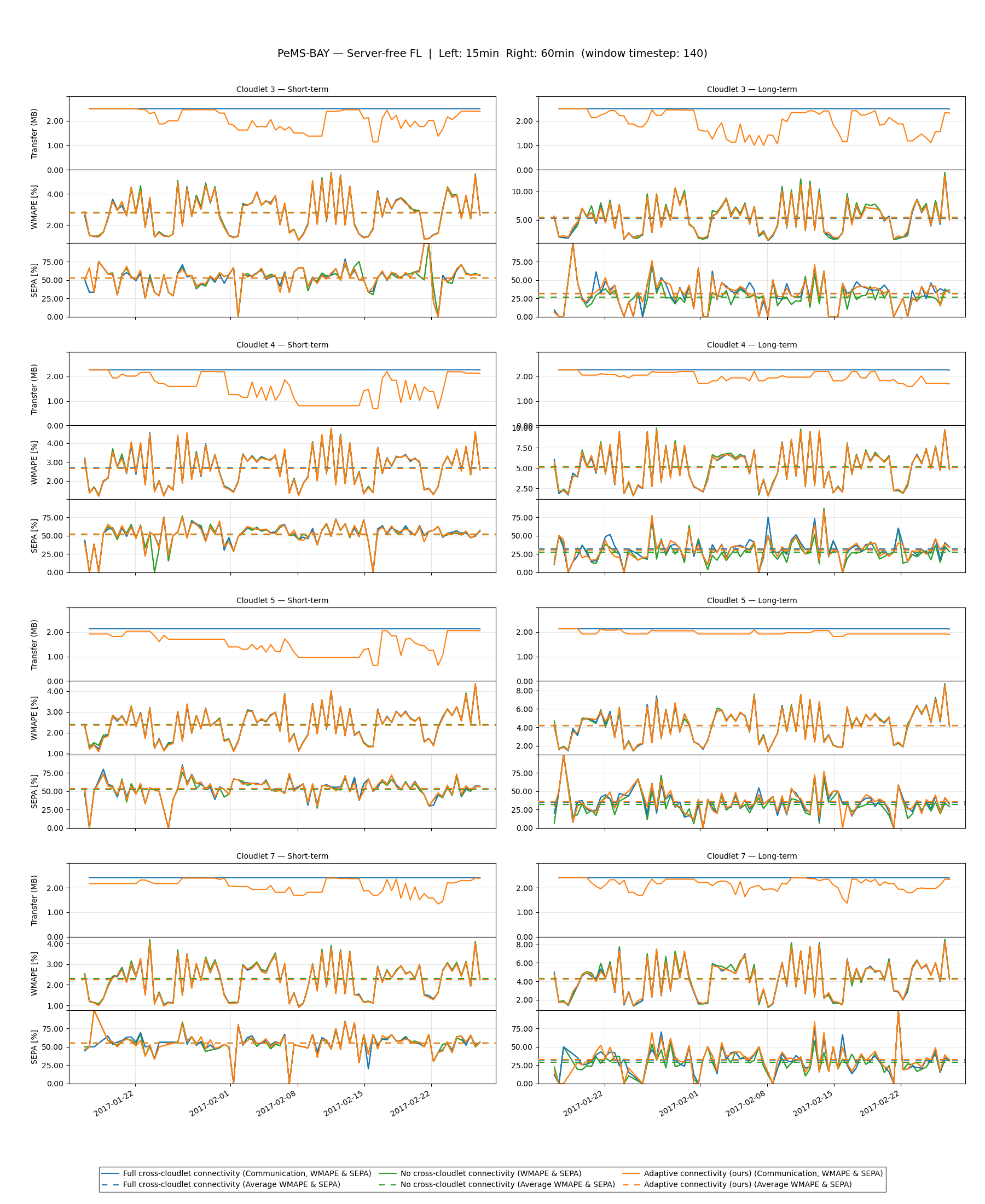}
  \caption{Communication cost, WMAPE, and SEPA for PeMS-BAY dataset and for four representative cloudlets (3, 4, 5, 7) under Server-free FL with 140-data timewindow (Left: short-term prediction; right: long-term prediction)}
  \label{fig:5}
\end{figure*}

\subsection{Impact on graph connectivity and data window sizes}
Table~\ref{table:4} presents results across all training setups, datasets, prediction horizons, and three connectivity regimes: (i) \emph{full cross-cloudlet connectivity} (cloudlets construct full ST-GNN subgraph by importing all needed $l$-hop neighbors, as seen in Fig.~\ref{fig:1}-d), (ii) \emph{no cross-cloudlet connectivity} (cloudlets use only local nodes and intra-cloudlet edges, as seen in Fig.~\ref{fig:1}-b), and (iii) \emph{adaptive connectivity} (selective removal of boundary nodes via the adaptive cross-cloudlet pruning algorithm).

When focusing on standard error metrics (MAE, RMSE, and WMAPE), the differences between setups are relatively minor across all prediction horizons. This observation is consistent with prior work showing that simplified spatial structures can perform comparably to models with full graph information \cite{experimental_results_1}. The reason RMSE (which penalizes large errors) is small across different connectivity strategies is due to how rare these sudden traffic events are. As shown in Table~\ref{table:3}, time steps classified as sudden events account for less than 0.3\% of the total validation samples in PeMS-BAY, with similar proportions observed in PeMSD7-M.

However, SEPA metric reveals new information that's not talked about in previous papers. The gap between different connectivity strategies becomes especially pronounced at longer horizons. While short-term predictions achieve similar SEPA regardless of connectivity, mid- and long-term predictions depend heavily on spatial information to correctly capture traffic jams and their recovery. Full cross-cloudlet connectivity and our proposed pruning algorithm consistently achieve noticeably higher SEPA values compared to the graph with no outside connections. These results indicate that traditional error metrics tend to obscure the benefits of graph connectivity, whereas SEPA exposes significant differences that directly reflect a model's ability to track rapid traffic dynamics. Importantly, pruning redundant cross-cloudlet information does not degrade performance; instead, it can improve predictive accuracy by preserving only the most relevant nodes.

Furthermore, increasing the size of the data window---i.e., the amount of new data processed per training epoch---from 70 to 140 data timewindow generally improves performance across both standard metrics and SEPA. Larger updates provide more complete temporal context per round, leading to more stable training and more accurate predictions, particularly in scenarios with frequent sudden traffic changes.

\begin{figure*}[ht!]
  \centering
  \includegraphics[width=1\linewidth]{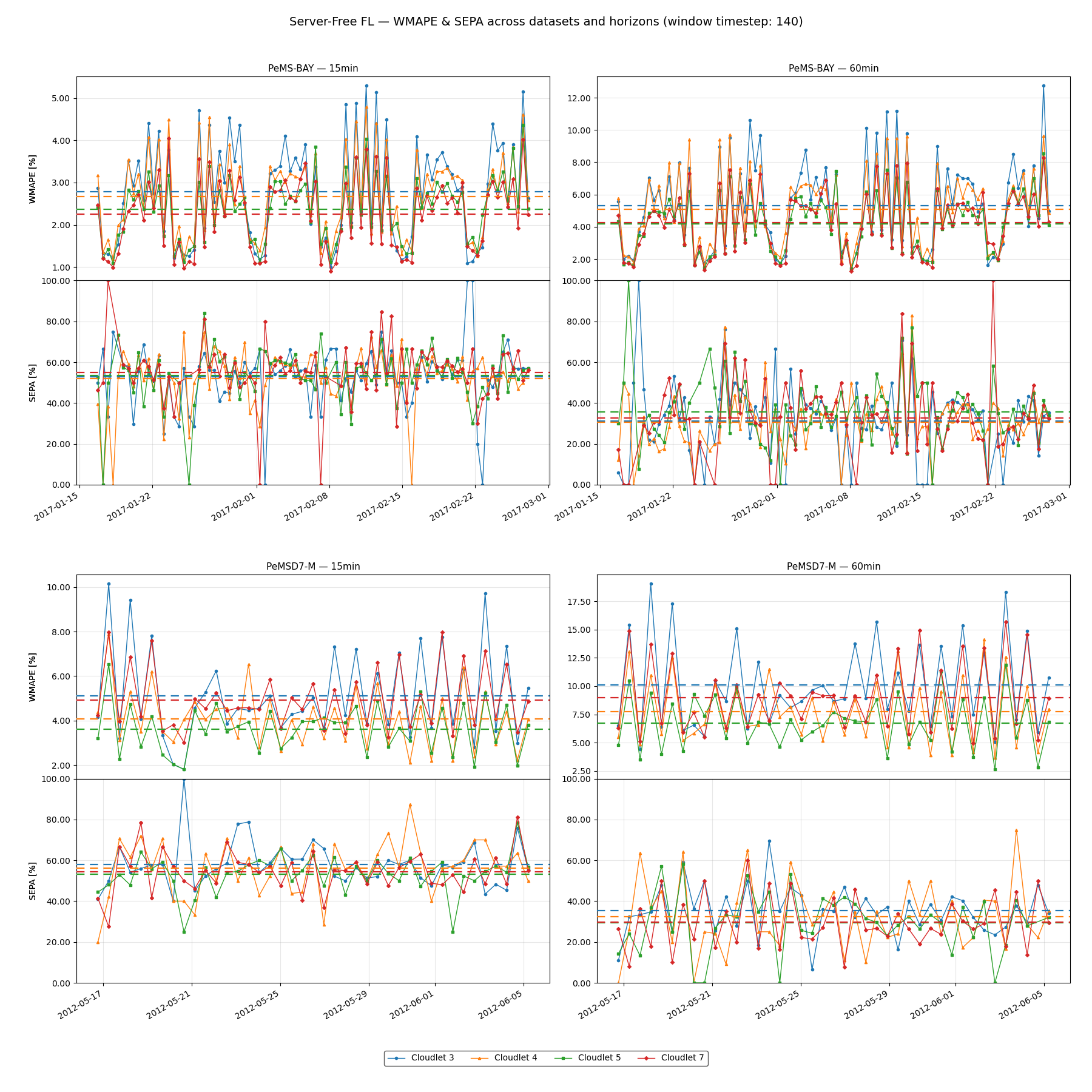}
  \caption{Cloudlet-level evaluation of WMAPE and SEPA for four representative cloudlets (3, 4, 5, 7) under Server-free FL with 140-data timewindow (Left: short-term prediction; right: long-term prediction) for adaptive connectivity}
  \label{fig:6}
\end{figure*}

\subsection{Effect of Graph Connectivity Pruning on SEPA}
Figure~\ref{fig:5} illustrates WMAPE and SEPA metrics, as well as communication cost across four representative cloudlets under the server-free FL setup on the PeMS-BAY dataset, evaluated for short-term (left side of the figure) and long-term (right side of the figure) predictions with 140-data timewindow. Each figure is divided into three rows per cloudlet, where the top row shows the communication cost of cross-cloudlet node features in megabytes, the middle row shows WMAPE in \%, and the bottom row shows SEPA in \% over the same time span.

In this subsection, we focus on analyzing the SEPA metric. Each subplot corresponds to a different cloudlet, showing how effectively sudden traffic events are captured under three connectivity conditions: full cross-cloudlet connectivity, no cross-cloudlet connectivity, and our proposed adaptive cross-cloudlet pruning algorithm. Only a subset of four cloudlets is presented here for readability, as their behavior is representative of the broader system trends observed across all cloudlets. Furthermore, to avoid misleading artifacts, time windows in which no sudden events occurred were excluded from the plots.

For short-term prediction (left side of the Figure~\ref{fig:5}), the results reveal almost no difference between the three approaches. Regardless of whether full cross-cloudlet connectivity, no cross-cloudlet connectivity, or pruning is applied, SEPA remains at similar levels across all cloudlets. This is further proven with horizontal reference lines representing the global average of SEPA metric across the entire set. We can see that horizontal lines yield similar SEPA performance. This aligns with the findings from Table~\ref{table:4}, where SEPA also showed only marginal differences in the short-term. Importantly, this behavior is consistent across different training setups, indicating that short-term accuracy is not significantly impacted by the choice of connectivity strategy.

As we extend the prediction horizon to long-term (right side of the Figure~\ref{fig:5}), the difference in correctly predicting sudden events changes across different graph connectivities as well. Full cross-cloudlet connectivity and the adaptive cross-cloudlet pruning algorithm consistently achieve higher SEPA values, capturing traffic jams and recoveries more effectively. In contrast, the no cross-cloudlet connectivity fails to detect many of these events, leading to visibly lower SEPA. This is further proven with the fact that horizontal line for no cross-cloudlet connectivity is lower compared to the full cross-cloudlet connectivity and our adaptive cross-cloudlet pruning algorithm. These results demonstrate that spatial context becomes increasingly important for longer-term predictions, and that selectively pruning less informative nodes preserves predictive quality while avoiding redundant communication.

We focus here on PeMS-BAY with 140-data timewindow, as these settings provide the clearest comparison between connectivity strategies. Additional results, including experiments with smaller data timewindow (70-data timewindow) and with the PeMSD7-M dataset, follow similar trends and are provided in the supplementary material for completeness.

\subsection{Analysis of adaptive connectivity impact on communication overhead}
Continuing from the previous subsection, we next analyze how our adaptive cross-cloudlet pruning algorithm modulates cross-cloudlet traffic over time and what this implies for communication cost, as shown in top and middle row of each representative cloudlet in Figure~\ref{fig:5}.

For communication cost subplots, we only compare full cross-cloudlet connectivity and the adaptive cross-cloudlet pruning algorithm. The no cross-cloudlet connectivity configuration is omitted from the communication plots because it involves no cross-cloudlet feature exchange, whereas we show full cross-cloudlet connectivity in order to emphasize the effect of our proposed cross-cloudlet pruning strategy. However, no cross-cloudlet connectivity configuration is included in the WMAPE plots to provide a reference for prediction quality without any inter-cloudlet communication.

Overall, our adaptive cross-cloudlet pruning algorithm achieves a substantial reduction in communication cost compared to full cross-cloudlet connectivity. We also show that communication cost does not decrease monotonically over time but instead fluctuates dynamically. This behavior stems from the adaptive nature of our algorithm: when sudden events are detected within a cloudlet, neighbouring cross-cloudlet nodes are temporarily protected from being removed in the current training window. Consequently, communication overhead can locally increase in regions experiencing frequent traffic dynamics, reflecting the model's adaptive retention of spatial context. At the same time, pruning reduces the size of the effective training subgraph, which lowers per-step computational load and shortens local training time. This reduction in processed data per iteration directly translates into lower energy consumption at cloudlets.

Despite the fluctuations in communication cost, the WMAPE curves remain nearly identical across all connectivity conditions for both short- and long-term forecasts. This is further proven with horizontal reference lines representing the global average of WMAPE metric across the entire set. We can see that horizontal lines yield similar SEPA performance for both prediction horizons. Although the variance slightly increases for long-term prediction, the overall differences in WMAPE between full cross-cloudlet connectivity, adaptive connectivity, and the no cross-cloudlet connectivity configurations are marginal. This indicates that conventional error metrics such as WMAPE are not sensitive enough to reflect the effects of connectivity pruning. In contrast, as demonstrated in the previous subsection, the SEPA metric more clearly exposes the performance trade-offs, particularly in long-term prediction,revealing where graph connectivity contributes most to capturing dynamic traffic behaviors.

As in the previous subsection, we focus on PeMS-BAY with 140-data time window. Additional results (70-data timewindow and PeMSD7-M) are provided in the supplementary material for completeness.

\subsection{Cloudlet Metric Analysis}
Figure~\ref{fig:6} presents cloudlet-level evaluation of the adaptive cross-cloudlet pruning algorithm under the server-free FL setup for both datasets with a 140-data timewindow, evaluating short-term (left) and long-term (right) predictions. Each figure is divided into two rows, where the top row shows WMAPE in \%, while the bottom row shows SEPA in \% over the same time span for all four representative cloudlets. WMAPE is highlighted as it provides a clear, percentage-based measure of relative error, making it easier to compare variability across cloudlets. Similar to the analysis conducted in our previous conference paper \cite{related_work_12}, the goal here is to examine how performance varies geographically across cloudlets and whether regional traffic patterns influence prediction quality in a online-training semi-decentralized environment.

The WMAPE (top rows) reveals consistent WMAPE spread across cloudlets, regardless of dataset or prediction horizon, indicating that performance variability stems from the datasets' geographical characteristics and sensor partitioning. Although the overall trend may appear visually uniform, the per-cloudlet horizontal WMAPE reference lines reveal clear differences in error magnitude, where each horizontal line represents global average of WMAPE metric across the entire set for each representative cloudlet. These disparities become even more pronounced in the PeMSD7-M dataset. This reinforces the fact that geographic heterogeneity and partitioning choices materially affect per-cloudlet accuracy, even in online-training scenario.

In contrast, the SEPA (bottom rows) show little to no spatial consistency and are considerably noisier across cloudlets. The horizontal SEPA reference lines shows slight differences in rate magnitude across individual cloudlets, where each horizontal line represents global average of SEPA metric across the entire set for each representative cloudlet. Furthermore, horizontal SEPA reference lines often diverge from the horizontal WMAPE reference lines, indicating that the ability to capture sudden traffic changes does not always align with traditional metrics. This discrepancy arises because traditional metrics measure average error across entire data timewindow, whereas SEPA explicitly quantifies the model's responsiveness to such events. 

For completeness, cloudlet-level figures for traditional FL and Gossip Learning (same datasets, horizons, and window size) are included in the supplementary material, where we can see that they display the same qualitative patterns as server-free FL.
\section{Related Work}
Graph Neural Networks have shown strong performance across various domains but often come with significant communication costs, especially when applied to large-scale graphs. This is further shown in distributed ST-GNNs due to large amount of duplicated node features that are transferred between cloudlets. Recent research has explored various strategies for reducing this overhead, ranging from graph or model sparsification in centralized training to parallelization and caching in distributed frameworks. However, most existing approaches either assume static graphs or full-batch training and are not designed for online training applications.

\subsection{Communication Reduction in Centralized training of GNNs}
% people are not message-passing to cut on communication, but for computation
In centralized GNN training settings, communication overhead is typically incurred during the initial data collection phase when node features from IoT-devices are transmitted to a central server. Once data is collected, the training process itself does not involve inter-device communication. As a result, there are currently two main approaches for reducing GNN training complexity by sparsification: \textit{simplifying the input graph} and \textit{sparsifying the model} \cite{related_work_11}.

The first approach utilizes sampling or pruning of the nodes or edges in the input graph and has been explored extensively. H. Peng et al. \cite{related_work_9} proposed a multi-stage graph neural network (m-GNN) to encode the topology of deep neural networks themselves as computational graphs. Their approach focuses on compressing CNN architectures such as ResNet and MobileNet by pruning channels under constraints such as FLOPs, using graph embeddings to drive policy learning. However, their approach is best suited for inference, not for training.

In another work, C. Chen et al. \cite{related_work_10} proposed a DyGNN framework where they prune nodes if they are considered as converged. The key idea is that once a node's representation has stabilized, it no longer needs to participate in the aggregation process, thus reducing redundant message exchanges. However, this technique is more suitable for static setups and not for online training, since in online training node features are updated continually with new streaming data, meaning it is difficult to determine in advance whether a node has converged. As a result, pruning decisions cannot be made reliably without the risk of degrading model performance.

On the other hand, model compression (or sparsification) is an under-explored area in the context of GNNs \cite{related_work_11}. However, in our prior work \cite{related_work_12}, we observed that model transfer sizes are negligible compared to the transfer size of node features. Hence, model sparsification does not significantly alleviate communication bottlenecks in our setting and will not be considered in this work.

\subsection{Communication Reduction in Distributed training of GNNs}
Distributed training of GNNs remains an active research area, with relatively few systems targeting GPU-based implementations \cite{related_work_4}. Unlike traditional distributed large-scale graph processing frameworks \cite{related_work_5, related_work_6}, distributed GNN training introduces greater communication challenges. This is primarily due to the layer-wise message passing required by GNN architectures. Each GCN layer or GAT layer requires sending/receiving features and gradients of neighbour vertices, where the dimension of vertex features and gradients is usually very large.

% difference over spatial-dimension vs time-dimension (check what they mean by full-batch and our mini-batch is on temporal-dimension)
Several works attempted to reduce communication overhead. One such work is the CAGNET framework \cite{related_work_2} introduces a family of communication-efficient algorithms, including 2D, 3D, and 1.5D parallel strategies. These methods restructure how node features, gradients, and model parameters are partitioned across devices, enabling a reduction in communication volume per iteration while maintaining training accuracy. While their approach is suited for full-batch training over spatial component, they provide no support for mini-batching over temporal component.

Similarly, the CDFGNN framework \cite{related_work_3} addresses communication bottlenecks by introducing a cache-based adaptive mechanism that stores historical vertex features and gradients locally. This reduces the frequency of remote accesses during training and is complemented by communication quantization techniques that compress messages between nodes. Just like CAGNET, CDFGNN framework has no support for mini-batch training over temporal component.

DistDGL \cite{related_work_7} framework reduces communication by partitioning the input graph using METIS \cite{related_work_8} to minimize edge cuts and co-locate data and computation. However, they don't address the issue regarding node feature transfer since their framework uses KVStore servers, meaning that mini-batch features are locally available via shared memory between machines, reducing the network traffic significantly. This significantly simplifies communication needs and is not representative of real-world distributed environments, where data must often be transferred over the network. 
Moreover, in real world, devices don't have a shared memory with other devices to exchange required node features.

\subsection{Communication reduction in ST-GNNs}
As noted in our previous work \cite{related_work_12}, distributed ST-GNNs incur significant communication overhead due to multiple node features being sent to multiple cloudlets.

Z. Jing et al. \cite{related_work_13} proposed TL-GPSTGN framework, which extends the ST-GCN model (also used in our experiments) with a Degree-based Pruning technique tailored for traffic prediction. Degree-based Pruning is used because it is more applicable for traffic prediction because the road network graph has a strong coupling relationship with the feature data. Their pruning strategy removes nodes deemed to have low influence on the prediction task. Specifically, they target nodes on the border of the road network for pruning, because they might depend on external, unmodeled influences, which their model cannot see. While their method achieves performance comparable to the base ST-GCN model, the pruning is applied statically, prior to the training phase. In contrast, our proposed algorithm performs dynamic node selection during the training phase. Rather than permanently removing nodes, our algorithm adaptively determines which node features are transmitted at each time step, allowing the cloudlets to retain visibility of the entire graph structure, while retaining prediction accuracy.

Another work by X. Liu et al. \cite{related_work_14}, further questions the efficiency of GNNs for traffic forecasting. Their SimST framework replaces graph message passing with lightweight spatial modules that pre-compute local ego-graphs and learn global spatial correlations through static node embeddings, avoiding runtime message exchange altogether. While this approach significantly reduces complexity, it effectively removes explicit graph connectivity during training. In our work, we demonstrate that graph connectivity benefits become evident during sudden events and long-term forecasts by introducing the SEPA metric.

To the best of our knowledge, there is currently no prior work focused on reducing communication overhead in distributed ST-GNNs during online training. Existing methods either focus on centralized settings, assume static datasets, or disregard the spatio-temporal dynamics inherent to traffic prediction tasks. Our work is the first to explore dynamic, per-timestep communication reduction for ST-GNNs in a distributed, cloudlet-based architecture under real-time constraints.

\subsection{Task-Specific Metrics for Traffic Prediction and Congestion Dynamics}
Most traffic prediction studies evaluate model performance using standard regression metrics such as MAE, RMSE, and (W)MAPE. While these metrics are useful for quantifying average point-wise error over time, several works have pointed out that they do not fully reflect the needs of traffic management systems, and should be complemented with application-specific measures \cite{related_work_15,related_work_16}.

A large body of work in traffic management system focuses on traffic congestion detection prediction \cite{related_work_17}. Many detection methods operate on multiple traffic variables simultaneously such as flow, density, occupancy, video-derived indicators, etc \cite{related_work_18}. G. R. Jagadeesh et al. \cite{related_work_19} moves closer to our setting by defining congestion based solely on vehicle speed time series. They compare the current speed at a sensor to a reference value such as its median free-flow speed, and declare the sensor congested if the current speed falls below a fraction $\alpha$ of that reference. Congestion labels are then refined by enforcing spatio-temporal consistency, for example by requiring that neighbouring sensors in space or time are also congested before confirming a congested state.

However, these methods are designed to predict whether a location is congested at a given time, which is easy to identify if vehicle speed gradually slows down. But traffic congestion can happen instantly, and is harder to detect. This abrupt shift from operation at free-flow conditions to congested conditions is called traffic flow breakdown and is typically the result of complex interactions in traffic dynamics. Existing work on flow breakdown prediction largely concentrates on identifying the onset of this transition, and typically does not explicitly quantify the accuracy of predicting recovery from breakdown back to free-flow conditions \cite{related_work_20}.

In contrast to the above, our work focuses on an event-centric evaluation of both congestion onset and recovery using only vehicle speed time series. Rather than asking whether a segment is congested at a given time, we explicitly detect sudden drops and rises in speed and evaluate how accurately a model predicts these event endpoints. To the best of our knowledge, there is no prior metric that is tailored to abrupt speed changes, jointly considers jam onset and recovery, and is designed to assess the behaviour of graph-based traffic predictors under different connectivity regimes.
\section{Future Work}
\subsection{Communication efficiency and representation pruning}
While the proposed adaptive cross-cloudlet pruning algorithm effectively reduces cross-cloudlet feature exchange, further improvements are possible by exploring more compact node representations and restricting the spatial receptive field of graph convolutions. One direction is to extend the current node feature pruning approach toward pruning node embeddings, or combining both strategies to further reduce redundant information exchange while preserving model expressiveness. Moreover, layer-wise GNN training schemes that limit the receptive field to a subset of neighbours \cite{future_work_1} could be adopted to minimize unnecessary feature propagation across distant nodes. Such hybrid pruning and receptive-field restriction mechanisms could substantially reduce bandwidth consumption.

\subsection{Cloudlet personalization}
Another promising direction is implementing individual cloudlet personalization to address performance variability across cloudlets. Our experiments revealed that prediction accuracy varies across cloudlets and across different time periods.

While our current approach performs cloudlet personalization by adapting each cloudlet's spatial context by pruning/protecting cross-cloudlet nodes, the underlying model is still shared and trained with a common objective. However, there are other personalized approaches to improve performance variability across cloudlets. Such a personalized approach may involve adjusting model parameters or incorporating local fine-tuning for cloudlets during time periods with sudden events detection rate, potentially enhancing prediction accuracy across the network while maintaining decentralized resilience.

\subsection{Cloudlet placement optimization}
An important limitation of our current approach is the predefined placement of cloudlets, and communication range, which has not been optimized for coverage or efficiency. The current cloudlet configuration is based on a static, distance-based partitioning of the sensor network. Future work should explore strategies for optimizing both the number and spatial placement of cloudlets to balance communication efficiency, coverage, and learning performance. Additionally, aligning cloudlet placement with real-world base-station infrastructure could improve practicality and facilitate deployment in existing smart mobility environments.
\section{Conclusion}
In this paper, we tackled the core communication overhead bottleneck of semi-decentralized ST-GNN training for traffic forecasting, which is induced by cross-cloudlet feature exchange. We introduced the Sudden Event Prediction Accuracy (SEPA), an event-focused metric that reveals graph connectivity truly matters during rapid traffic changes, and proposed an adaptive cross-cloudlet pruning algorithm, a boundary-aware pruning policy that adaptively reduces cross-cloudlet nodes for each individual cloudlet based on their performance.

Experimental results on the PeMS-BAY and PeMSD7-M datasets show that standard validation metrics remain close across all connectivity setups, semi-decentralized setups and prediction horizons, yet these metrics mask the benefits of graph connectivity, making ST-GNNs appear inefficient. Our new metric reveals advantages of full cross-cloudlet connectivity and of our pruning algorithm compared to no cross-cloudlet connectivity, especially at longer horizons. The adaptive cross-cloudlet pruning algorithm lowers per-cloudlet communication while maintaining competitive predictive accuracy, demonstrating that communication can be trimmed without eroding event-detection reliability.

\section*{Acknowledgment}
This work has been supported in part by the Horizon Europe WIDERA program under the grant agreement No. 101079214 (AIoTwin) and by the European Regional Development Fund under grant agreement PK.1.1.10.0007 (DATACROSS).

\bibliographystyle{IEEEtran}
\bibliography{literature}

\begin{IEEEbiography}[{\includegraphics[width=1in,height=1.25in,clip,keepaspectratio]{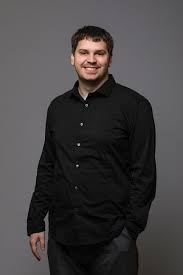}}]{Ivan Kralj} is a Ph.D. student at the Faculty of Electrical Engineering and Computing (FER), University of Zagreb, Croatia. He received the B.Sc. and M.Sc. degrees in Information and Communication Technology from FER in 2018 and 2020, respectively. Since 2020, he has been a researcher and assistant at the Department of Telecommunications at FER, where he participates in subjects such as distributed systems and Internet of Things. He is involved in the EU Horizon project AIoTwin and currently conducts research on distributed and semi-decentralized training of graph neural networks for edge-enabled Internet of Things systems.
\end{IEEEbiography}

\begin{IEEEbiography}[{\includegraphics[width=1in,height=1.25in,clip,keepaspectratio]{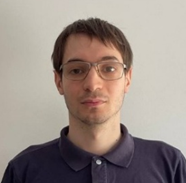}}]{Lodovico Giaretta} is currently Senior Researcher at RISE Research Institutes of Sweden, where he participates in or leads several European, national and industrial research projects. He previously obtained his PhD degree from KTH Royal Institute of Technology, Sweden, in 2023, with a thesis titled "Towards Decentralized Graph Learning". His research interests lay at the intersection of several areas, including large-scale decentralized deep learning, graph representation learning, privacy-preserving AI and decentralized AI agents.
\end{IEEEbiography}

\begin{IEEEbiography}[{\includegraphics[width=1in,height=1.25in,clip,keepaspectratio]{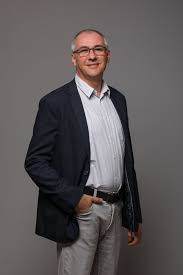}}]{Gordan Ježić} is a Full Professor with the Department of Telecommunications, Faculty of Electrical Engineering and Computing (FER), University of Zagreb, Croatia. He received the Ph.D. degree in Electrical Engineering from FER in 2003. He has been affiliated with the Department of Telecommunications since 1996 and served as the Head of the Department from 2016 to 2020. His research interests include next-generation communication networks and services, machine-to-machine communication in Internet of Things environments, and software technologies in telecommunications, with a focus on communication protocols and multi-agent systems. He has participated in numerous national and international research projects and has long-standing collaboration with industry partners. He has co-authored over 100 scientific publications and has served on editorial boards and program committees of several international journals and conferences.
\end{IEEEbiography}

\begin{IEEEbiography}[{\includegraphics[width=1in,height=1.25in,clip,keepaspectratio]{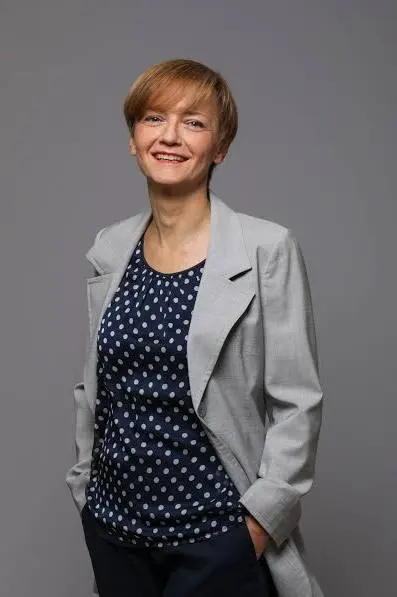}}]{Ivana Podnar Žarko} is a Full Professor at the Faculty of Electrical Engineering and Computing (FER), University of Zagreb, Croatia, where she teaches courses on distributed systems and the Internet of Things. She received the B.Sc., M.Sc., and Ph.D. degrees in Electrical Engineering from FER in 1996, 1999, and 2004, respectively. She leads FER's Internet of Things Laboratory and currently coordinates the Horizon Europe project AIoTwin. She has participated in numerous national and European research projects, including serving as Technical Manager of the H2020 project symbIoTe. Her research interests include large-scale distributed systems, IoT interoperability, and distributed ledger technologies. She has co-authored over one hundred scientific publications and has served on program committees and editorial boards of several international conferences and journals.
\end{IEEEbiography}

\begin{IEEEbiography}[{\includegraphics[width=1in,height=1.25in,clip,keepaspectratio]{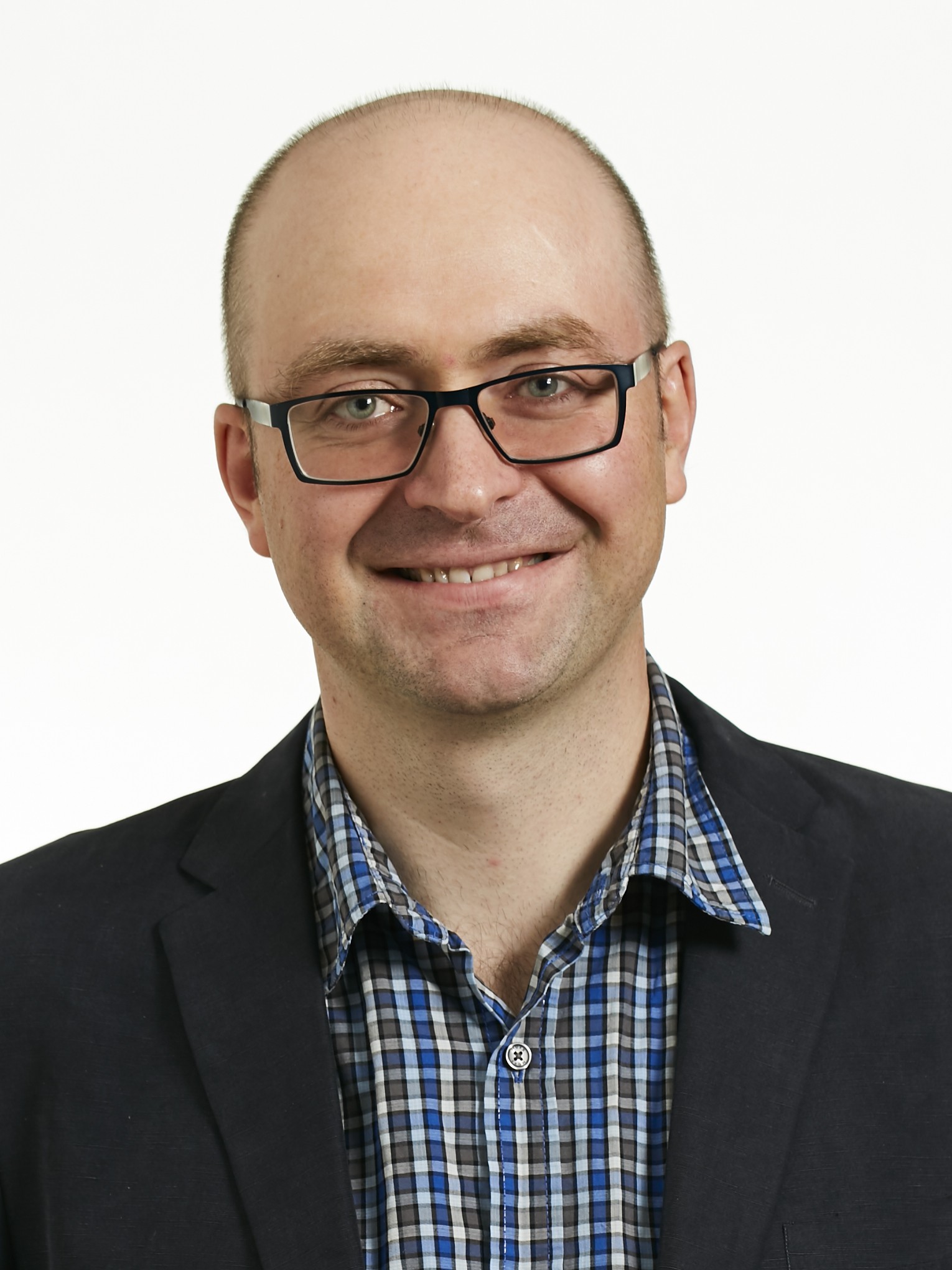}}]{\v{S}ar\=unas Girdzijauskas} received the PhD degree from École Polytechnique Fédérale de Lausanne (EPFL), Switzerland, in 2009. He was appointed Docent in Computer Science at KTH Royal Institute of Technology, Stockholm, Sweden, in 2016. He is currently a Full Professor with the Computing and Learning Systems Department at KTH, where he also serves as Vice Head of Department. He is also a Senior Researcher at RISE Research Institutes of Sweden. He has graduated ten PhD students and has coordinated two EU Marie Curie Initial Training Network projects. His research interests lie at the intersection of distributed systems and machine learning, with a particular focus on graph learning and privacy-preserving federated and decentralized algorithms.
\end{IEEEbiography}

\EOD

\end{document}